\def\FORMAT{0} 

\ifnum\FORMAT=0 
    \def\ARXIV{1} 
    \def\PAGENUMS{1} 
    \def\TOC{1} 
    \def\BLIND{0} 
    \def\BBL{0} 
\else
    \ifnum\FORMAT=1 
        \def\ARXIV{0} 
        \def\PAGENUMS{1} 
        \def\TOC{0} 
        \def\BLIND{1} 
        \def\BBL{0} 
    \else
        \def\ARXIV{0} 
        \def\PAGENUMS{0} 
        \def\TOC{0} 
        \def\BLIND{0} 
        \def\BBL{1} 
    \fi
\fi

\ifnum\BLIND=0
    \documentclass[authoryear,preprint,12pt]{elsarticle}
\else
    \documentclass[authoryear,doubleblind,preprint,review,12pt]{elsarticle}
    \usepackage{lineno}
\fi

\usepackage{amssymb}

\usepackage{booktabs}
\usepackage[hyphens]{url}
\usepackage[hidelinks,breaklinks]{hyperref}
\usepackage{multirow}
\usepackage{xcolor}
\usepackage{threeparttable} 
\usepackage{siunitx} 
\usepackage[normalem]{ulem} 
\robustify\uline
\usepackage{subcaption}
\usepackage{orcidlink}

\usepackage{graphicx}
\usepackage{amsmath}
\usepackage{tcolorbox}
\usepackage{xparse}
\usepackage{soul}
\usepackage{longtable}
\useunder{\uline}{\ul}{}
\usepackage{float}
\usepackage[section]{placeins}  
\ifnum\ARXIV=0
    \journal{International Journal of Forecasting}
\else
    
    \usepackage{fancyhdr}
    \usepackage{lastpage}
    \fancypagestyle{plain}{
      \fancyhf{}
      \fancyhead[C]{\footnotesize 
        Accepted manuscript: 
        Clegg and Cartlidge (2027), 
        International Journal of Forecasting
        }
      \fancyfoot[C]{\footnotesize
        Page \thepage\ of \pageref{LastPage}}

    }

    \fancypagestyle{pprintTitle}{
    \fancyhf{}
    \fancyfoot[L]{\footnotesize
        Accepted author manuscript,
        International Journal of Forecasting (2027)}
    \fancyfoot[R]{\footnotesize Page \thepage\ of \pageref{LastPage}}

    }
\fi

\usepackage{xspace}
\usepackage[color=orange9]{todonotes}

\definecolor{verbosegray}{gray}{0.975}

\begin{document}

\ifnum\ARXIV=1
    \pagestyle{plain}  
\fi

\pagestyle{plain}

\begin{frontmatter}


\title{Capturing Intransitive Dominance in Tennis Forecasting: A Graph Neural Network Approach}

\ifnum\BLIND=1
    \author{Anonymous Author(s)} 
\else
    \author{Lawrence Clegg\corref{cor1}\,\orcidlink{0009-0009-1292-4773}}
    \ead{lawrence.clegg@bristol.ac.uk}
    \author{John Cartlidge\,\orcidlink{0000-0002-3143-6355}}
    \ead{john.cartlidge@bristol.ac.uk}
    \cortext[cor1]{Corresponding author}
    \affiliation{
            organization={School of Engineering Mathematics and Technology, University of Bristol},
            addressline={\\Ada Lovelace Building, Tankard's Close},
            city={Bristol},
            postcode={BS8 1TW}, 
            country={UK}}
    \renewcommand{\theaffn}{} 
\fi

\begin{abstract} 
Intransitive player dominance, where player A beats B, B beats C, but C beats A, is common in competitive tennis. Yet, there are few known attempts to incorporate it within forecasting methods. We address this problem with a graph neural network approach that explicitly models these intransitive relationships through temporal directed graphs, with players as nodes and their historical match outcomes as directed edges. Our model (65.7\% accuracy, 0.214 Brier score) forecasts competitively with established rating systems such as Weighted Elo. Although it does not improve on the baseline in unconditional accuracy, a forecast-encompassing test shows that it carries complementary information. A combined forecast significantly outperforms Weighted Elo, and there is some indication that the gain is larger on the intransitive matchups our model targets. A graph-based representation of player interactions thus captures a forecasting signal that transitive rating systems discard, even between players who share no common opponents.
\end{abstract}

\ifnum\ARXIV=1
\else
    \begin{highlights}
        \item Research highlight 1 - We develop a novel tennis match outcome forecasting model using temporal directed graphs encoding player matchup histories and a spectral graph convolutional network that learns from intransitive player relationships.
        
        \item Research highlight 2 - Our model predicts match outcomes with 65.7\% accuracy and a Brier score of 0.214 across both men's and women's tours, performing competitively with established rating systems such as Weighted Elo.

        \item Research highlight 3 - A forecast-encompassing test shows our model carries information that a strong rating-system baseline lacks: a combined forecast significantly improves on the baseline.

        \item Research highlight 4 - We introduce two measures of intransitivity and find cyclic dominance is prevalent in professional tennis, more so in the women's game (12.1\% more intransitive triads).
        
    \end{highlights}
\fi


\end{frontmatter}


\ifnum\PAGENUMS=1
    \pagenumbering{arabic} 
\else 
    \pagenumbering{gobble} 
\fi

\newpage

\ifnum\TOC=1
    \tableofcontents
    \newpage
\fi

\ifnum\BLIND=1
    \linenumbers  
\fi

\section{Introduction}\label{sec:intro}
\noindent
Tennis is a sport well-suited to predictive modelling, with dense tournament schedules generating extensive historical data. The official ranking systems of the Association of Tennis Professionals (ATP) and Women's Tennis Association (WTA) have been shown to exhibit some predictive power for match outcomes \citep{clarkedyte2000,klaassen2003forecasting}, but there are notable limitations: for example, ranking points accumulate over a 52-week period, without decay, which can mask recent changes in player form, while match-specific factors, such as surface type, tournament progression difficulty, and margin of victory in individual matches, are overlooked. 

Some well-known methods have been applied to tennis and modified to accommodate these factors, such as a Bradley-Terry model with surface-specific adjustments \citep{mchale2011bradley} or Elo rating systems that incorporate margin of victory \citep{kovalchik2020extension, angelini2022weighted}.
More recently, analysis has extended to progressively finer scales: from match outcomes to point-level dynamics such as unforced errors \citep{peiris2025unforced}, rally-level player characteristics \citep{epasinghege2025rally}, and shot-level tactics such as intended serve direction \citep{tea2023analysis}.

Bookmakers are considered the most accurate predictors of match outcomes \citep{kovalchik2016searching}, with sophisticated models that adjust odds based on betting patterns and proprietary methods. Yet, despite the multi-billion dollar betting industry, one limitation that persists is the poor consideration of intransitivity \citep{van2025non}. Intransitivity is analogous to rock-paper-scissors. In tennis, it occurs where player A tends to defeat B, B defeats C, yet C defeats A, violating the assumption of transitive dominance. One example is Nadal beats Federer (24/40 career win rate), Federer beats Davydenko (19/21), yet Davydenko beats Nadal (6/11). Such patterns are known to be common in tennis, with two papers finding evidence in both the WTA and ATP tours \citep{bozoki2016application, temesi2019interactive}, and recent work has shown these patterns create systematic biases in rating systems like Elo \citep{hamilton2024elo}.

Representing historical tennis matches as a graph (where each player-matchup can be represented as an edge joining two player-nodes) can preserve intransitive dynamics. Several existing graph-based approaches have reported strong predictive performance, such as the eigenvector centrality approach by \cite{Arcagni2023}. However, global ranking methods such as eigenvector centrality cannot capture intransitive matchups, where player-specific dynamics matter more than overall strength. To better handle these scenarios, a graph model must effectively learn from the local neighbourhoods of players. 

We tackle this fundamental limitation of sports forecasting models with graph neural networks (GNNs), 
\ifnum\BLIND=1{a predictive approach that has not previously been applied to pre-match tennis prediction}\else{a predictive approach that has not previously been applied to pre-match tennis prediction \citep[apart from our own preliminary study:][]{clegg2025gnn}}\fi.\footnote{A search was conducted on 22/10/2025 using Google Scholar to identify papers applying GNNs to tennis prediction. The search combined six graph neural network related terms with eleven tennis prediction related terms, resulting in 66 distinct search queries. All searches used exact phrase matching with quotation marks (e.g., ``graph convolutional network'' ``tennis forecasting''). In total, these search queries returned 11 papers. Other than our own preliminary work \ifnum\BLIND=0
  \citep{clegg2025gnn},
\else
  [Anonymous, 2025b],
\fi none of the papers used GNNs for pre-match tennis outcome forecasting.}
We first construct gender-specific surface graphs that summarise match history and preserve player relationships, and apply a spectral graph convolutional network for directed graphs named MagNet \citep{zhang2021magnet}. We then establish that our model performs competitively with established rating systems such as the Weighted Elo approach by \cite{angelini2022weighted}, although it trails the Pinnacle Sports bookmaker benchmark. Next, we analyse model behaviour on intransitive matchups, where intransitivity is measured using a model-free count of cyclic head-to-head triads and an evidence-weighted adaptation of the graph-based measure of \citet{hamilton2024elo}: we find that women's tennis contains 12.1\% more cyclic triads than men's. Finally, we use a forecast-encompassing test to show that Weighted Elo does not fully subsume our model, whose marginal contribution remains significant on intransitive matchups.

In Section~\ref{sec:background}, we describe the related background of tennis forecasting, graph-based predictive methods, and intransitivity. In Section~\ref{sec:data}, we describe the necessary data for our research. Section~\ref{sec:method} introduces our graph-based method, while in Section~\ref{sec:results}, we report the models' out-of-sample accuracy. In Section~\ref{sec:discussion}, we analyse intransitivity and test whether our model carries forecasting information complementary to established rating systems. Finally, in Section~\ref{sec:conclusion}, we summarise our findings, acknowledge key limitations, and outline future research directions.

\section{Background}\label{sec:background}
\noindent
Tennis presents an ideal domain for predictive modelling due to several of its unique characteristics. Its scoring structure is inherently hierarchical, advancing from points to games to sets, which lends itself to mathematical modelling. Furthermore, the individual nature of singles tennis eliminates the complexity of team dynamics found in other sports, where player transfers and injuries can confound prediction models. The worldwide tournament calendar also ensures that the same players face one another repeatedly on multiple surfaces and across various event tiers, thereby generating a dense record of head-to-head encounters that can be analysed for statistical inference.

Various predictive modelling approaches have been developed for tennis. Point-based models determine match win probabilities by first estimating on-serve and return strengths to obtain point win probabilities and then applying hierarchical equations that reflect the structure of tennis scoring; examples include the Bayesian model of \citet{Ingram2019} and the serve-performance forecasts of \citet{Gollub2021}. In contrast, pairwise comparison models use head-to-head match outcomes to dynamically update relative player strength scalars, as in the popular Elo rating system~\citep{elo1978rating}. Regression models using player statistics as predictors can also be used; \citet{Buhamra2024} compared several such specifications for Grand Slam matches. \cite{kovalchik2016searching} compared the performance of various published methods across these categories, finding that a specialised Elo implementation by \cite{FiveThirtyEightSerenaGOAT} achieved 70\% accuracy, second only to the Bookmaker Consensus Model (BCM) of \cite{leitner2009federer} at 72\%. \citet{Wilkens2021} similarly find that machine-learning models plateau near 70\% accuracy, with most predictive information already embedded in betting-market odds.

While the original Elo rating model by \cite{elo1978rating} was intended for chess, its application to tennis is well-established. \cite{kovalchik2020extension} developed four distinct models that extended Elo by incorporating various margin of victory measures, including: set scores, games won, break points won, total points won, and serve percentage won. \cite{angelini2022weighted} proposed a similar method, named Weighted Elo, that adjusts rating updates based on the proportion of games won in each match. Both methods demonstrated improvements over the standard Elo system, confirming that incorporating greater match information leads to more accurate predictions.

One group of models that the survey by \cite{kovalchik2016searching} did not consider was graph-theoretic methods.
Two early studies that applied graph theory to tennis offered alternative rating systems to the official ATP ratings. \cite{Radicchi2011} applied PageRank, a node importance algorithm originally developed by \cite{pagerank}, to identify the most successful men's tennis players, using a dataset of matches played from January 1968 to October 2010. They observed that their ranking method outperformed the ATP and ITF rankings by anticipating, a year in advance, the player who would top those official rankings. However, they found active players were penalised compared to retired ones due to their incomplete career data. Subsequently, \citet{DingleKnottenbeltSpanias2013} extended Radicchi's PageRank approach by evaluating it as a match-outcome predictor, finding it slightly outperformed official rankings on ATP matches (67.0\% vs 66.4\%). However, the authors did not distinguish between court surfaces, and lower-ranked players appeared in too few matches to be reliably ranked.

\cite{Aparico2016} incorporated court surface considerations with a coloured directed network. With the use of 4-node subgraphs, the authors developed an Orbit-score ranking system that captured indirect dominance relationships between players from different time periods. The model identified surface specialists, all-round players, and players with an Achilles-heel on a specific surface. While their model enabled more detailed player differentiation, it was not extended to directly predict individual match outcomes.
More recently, \cite{Bayram2021} used three centrality measures (out-in-degree-difference, Hubs, and PageRank) to estimate surface-specific player scores. The score was used as a feature in several machine learning models to predict individual matches, with a random forest model achieving the highest classification accuracy of 67\% on a dataset of 21,083 matches between 2012 and 2020. The authors' approach demonstrated the effectiveness of graph-based features, but prioritised machine learning sophistication over the graph methods themselves.

\cite{Arcagni2023} developed a new model based on the eigenvector centrality, deriving global ratings that were subsequently used as covariates in a simple logit model for match prediction. The eigenvector centrality approach attained a Brier score of 0.194 on ATP matches between 2016 and 2020, outperforming several competing methods, including standard Elo (0.206) and the margin of victory Elo extension (0.204) by \cite{kovalchik2020extension}. This impressive performance demonstrated the growing effectiveness of graph-theoretic methods in tennis prediction while suggesting potential for even more sophisticated approaches.

In other domains, recent advances in deep learning applied to graphs, particularly Graph Neural Networks (GNNs), have shown considerable promise. There have been successful applications of GNNs in other sports, such as American football and Counter-Strike: Global Offensive \citep{xenopoulos2021graph}, association football \citep{mirzaei2022sports}, and basketball \citep{he2022gnnrank}.

Unlike traditional GNN architectures designed for undirected or symmetric graphs, recent work on spectral methods for directed graphs enables learning from cyclic structures through complex-valued representations. \cite{zhang2021magnet} used the magnetic Laplacian (a complex Hermitian matrix) to encode directed cycle information in its eigenvalue spectrum, with a tunable parameter $q$ that controls sensitivity to cyclical patterns.
Such spectral methods can address intransitive relationships between nodes, where $A \to B \to C \to A$ forms a directed cycle that cannot be resolved into a strict hierarchical ordering. The capability to learn from directed cycles, rather than through symmetrisation, makes spectral methods for directed graphs specifically suited to tennis, where intransitive matchup patterns are prevalent \citep{bozoki2016application, temesi2019interactive}. Representing tennis players as nodes and their matchup histories as directed edges, GNN architectures like MagNet \citep{zhang2021magnet} can encode intransitive matchup patterns as directed-cycle information that scalar ratings cannot represent.

\section{Data}\label{sec:data}
\noindent
We collected historical data on professional tennis matches organised by both the Association of Tennis Professionals (ATP) and the Women's Tennis Association (WTA) from \href{http://www.tennis-data.co.uk}{tennis-data.co.uk}, spanning 8 January 2012 to 16 October 2025.

The dataset contains information such as match date, competitors, games won by each competitor for each set played, tournament, surface, round of the tournament, and betting odds from several bookmakers, from which we select Pinnacle Sports due to the consistent availability of their odds across the dataset timespan. Additionally, Pinnacle Sports is recognised as a ``sharp'' bookmaker, one that operates with low margins and adjusts odds aggressively based on large betting volumes, which leads to accurate outcome forecasting \citep{buchdahl-squares-and-sharps,hegarty-ijf-2025}. We also gathered the following player characteristics from \href{https://www.tennisexplorer.com/}{tennisexplorer.com}: height, weight, date of birth, and handedness. For missing characteristic values, the median of the population is imputed. These static attributes are evaluated as candidate node features but do not improve predictive accuracy, and are excluded from the final model (Section~\ref{subsec:ablation}).

We filter for matches played in the highest tiers of professional tennis for both men and women: Grand Slams, the season-ending tour finals, and all 1000 and 500 series events. We treat men's and women's tennis as separate prediction tasks, training and validating each tour independently; only hyperparameter tuning is run jointly, so both tours share a single configuration.
 
After removing incomplete matches and entries with missing or incorrect values, the final dataset comprises 20,181 matches played by 670 men and 19,985 matches played by 640 women. We split each tour chronologically into three consecutive sets. The earliest matches build the dominance graph and train the model. The next 8,435 matches form a validation set, used for hyperparameter optimisation (Section~\ref{subsec:paramopt}) and the ablation study (Section~\ref{subsec:ablation}). The most recent 12,052 matches are held out as the out-of-sample test set. The men's validation set runs from 31 July 2018 to 21 February 2022 and the women's from 28 August 2018 to 12 March 2022, with each test set beginning the next day and ending in October 2025. The walk-forward procedure is detailed in Section~\ref{subsec:walkfwdval}.

\section{Graph Model}\label{sec:method}
\noindent
In their overview of temporal graph networks, \cite{longa2023} describe two primary approaches to temporal graph representation: snapshot-based (using a sequence of static graphs capturing network states at different time points) and event-based (representing edges as a continuous stream of timestamped interactions). Here, we adopt a snapshot-based approach by interpreting each tournament round (e.g., Wimbledon 2010 Quarter Finals) as a discrete timestamped snapshot. 

\begin{table}[tb]
\centering
\caption{Mathematical notation, grouped by role.}
\label{tab:notation}
\scriptsize
\begin{tabular}{@{}cl@{}}
\toprule
\textbf{Symbol} & \textbf{Description} \\
\midrule
\multicolumn{2}{@{}l}{\textit{Indices}} \\
$s$              & Target court surface for which the graph is built (clay, grass, or hard) \\
$s_k$            & Court surface of historical match $k$ \\
$i, n$           & Snapshot indices: arbitrary ($i$), prediction target ($n$) \\
$k$              & Historical match index \\
$u, v$           & Player indices \\
\addlinespace
\multicolumn{2}{@{}l}{\textit{Observed data and graph structure}} \\
$\tau_n, \tau_k$ & Timestamps: snapshot $n$ (earliest scheduled match); match $k$ \\
$g_k(u,v)$       & Proportion of games won by $u$ against $v$ in match $k$ \\
$V$              & Node set (players) \\
$E_i$            & Edge set at snapshot $i$ (head-to-head matchups) \\
$G^s_i$          & Surface-specific graph at snapshot $i$ \\
$\mathbf{X}^V_i$ & Node feature matrix at snapshot $i$ \\
\addlinespace
\multicolumn{2}{@{}l}{\textit{Optimised parameters}} \\
$\lambda$        & Time decay rate \\
$\alpha_{s,s_k}$ & Surface similarity (skill transfer from $s_k$ to $s$) \\
$\beta_k$        & Tournament prestige (by tier of match $k$) \\
$q$              & Directional weighting in the magnetic Laplacian (phase strength) \\
\addlinespace
\multicolumn{2}{@{}l}{\textit{Derived quantities and estimators}} \\
$\phi_k$         & Time decay coefficient for match $k$ (from $\lambda$) \\
$W^s_i, w_{uv}$  & Edge weights for surface $s$; weight from $u$ to $v$ \\
$D^s_n(u,v)$     & Dominance score of $u$ against $v$ on surface $s$ at snapshot $n$ \\
$\hat{p}_{uv}$   & Estimated set-win probability, $u$ against $v$ \\
$\hat{P}_3, \hat{P}_5$ & Estimated match-win probabilities (best-of-3, best-of-5) \\
\bottomrule
\end{tabular}
\end{table}

We now describe our graph construction, the use of MagNet for match prediction, and our walk-forward validation approach. Table~\ref{tab:notation} summarises the key mathematical notation used in this section.

\subsection{Graph representation}\label{subsec:graphrepresentation}
\noindent
In professional tennis, players compete on three distinct court surfaces: clay, grass, and hard. Each surface requires different playing styles and favours different skill sets, with researchers such as \cite{Fayomi2022} finding that the ``surface on which a game is played on contributes significantly towards a player's performance''. Players compete separately on the men's tour (ATP) and women's tour (WTA), with no cross-gender competition. We construct separate graph representations for each gender and each surface, accounting for this well-documented variation in player performance across surfaces, although we incorporate cross-surface information through our dominance score calculation, which aggregates match results from all surfaces with appropriate weighting.

A static directed graph can be formally defined as $G=(V,E,\mathbf{X}^V, \mathbf{X}^E)$, where $V$ represents a set of nodes (or vertices), $E$ represents the set of edges between nodes, $\mathbf{X}^V$ represents the associated node features, and $\mathbf{X}^E$ represents the associated edge features. The edges are directed, meaning $(u,v)$ does not necessitate $(v,u)$ and $E$ is asymmetric. Importantly, each edge $(u,v)$ can possess a weight $w_{uv} \in \mathbf{X}^E$ that quantifies the strength of the relationship.

For each tournament round snapshot, we construct three surface-specific graphs per gender, summarising all the preceding matches. By encoding surface in the graph, we aim to model surface-dependent performance dynamics more precisely. Hence, for each snapshot $i$ and court surface $s$ (clay, grass, or hard), we construct the graph:
\begin{equation}\label{eq:graph}
G^s_{i} = (V, E_{i}, \mathbf{X}^{V}_{i},  W^s_i),
\end{equation}
where $W^s_i$ denotes the edge-weight vector. Since this is the only edge feature we consider, we use the simplified notation $W^s_i$ rather than the more general edge feature matrix $\mathbf{X}^{E,s}_{i}$.

We consider the set of all players of a given gender who competed during the period covered by the dataset as the fixed node set $V$. The node features $\mathbf{X}^V_i \in \mathbb{R}^{|V|\times d}$ are the surface-specific in-degree and out-degree of each player, summarising the number of incoming and outgoing dominance edges for each player-node. We also evaluated static player attributes (height, weight, date of birth, and handedness) as node features, but they did not improve predictive accuracy and are excluded from the final model (Section~\ref{subsec:ablation}). Each feature vector \(\mathbf{v}\) is $\ell_2$-normalised, scaling the vector by its Euclidean norm \(\|\mathbf{v}\|_2 = \sqrt{\sum_j v_j^2}\), resulting in unit-length vectors that preserve directional information while standardising magnitude. Maintaining the fixed player set $V$ across surface-specific graphs allows the model to process the surface-specific degree features consistently across surfaces.

We use a common edge index set $E_i \subseteq V \times V$ across all surfaces, summarising historical head-to-head matchups between players, with edge weights $W^s_i$ that are surface-specific. Provided there is at least one match played between player $u$ and player $v$ prior to snapshot $i$, we add a weighted directed edge $(u,v)$ to $E_i$. Each weight \( w_{uv} \in W^s_i \) represents the head-to-head dominance of player $u$ over player $v$, where direction indicates the dominant player and magnitude quantifies the degree of dominance. The edge weight $w_{uv}$ is calculated based on the historical proportion of games won by player $u$ against player $v$. This base value is then adjusted using factors for time decay, surface weighting, and tournament weighting. Motivated by \citet{kovalchik2020extension}, who found granular performance measures more robust than binary match outcomes, we treat the dominance signal, the proportion of sets, games, or points won, as a tuned hyperparameter (Table~\ref{tab:paramtuning}). The proportion of games won $g_k(u,v) \in [0,1]$ by player $u$ against $v$ in historical match $k$ achieved the best validation Brier score on both tours, so we adopt it as the primary variable in our dominance score calculation.

An early use of exponential time decay to down-weight older sports match results originates from \cite{dixon1997modelling} for association football, where they down-weighted each historical match in the likelihood by an exponential factor. It was subsequently adapted to tennis by \cite{mchale2011bradley}, who additionally incorporated surface weights to improve forecasting accuracy. We further extend this principle by introducing an explicit tournament quality factor, so that results from higher-prestige events have greater influence on the estimated dominance between players. Hence, to each historical result $g_k(u,v)$ we apply the time decay coefficient:
\begin{equation}
    \phi_k(u,v) = \exp(-\lambda\,(\tau_n - \tau_k)),
\end{equation}
\noindent
where $\tau_n$ denotes the timestamp of the earliest scheduled match in the tournament round being predicted, $\tau_k$ denotes the timestamp of historical match $k$, and the decay rate $\lambda > 0$ is a learnable parameter governing the rate at which older matches lose influence. This approach ensures that the influence of past interactions diminishes over time, with more recent interactions having a greater impact on the current dominance scores.

For each target surface $s$, we calculate surface-specific dominance scores that incorporate matches from all surfaces with appropriate weighting. To refine the influence of each historical match, we scale each match by two further coefficients: a surface similarity parameter $\alpha_{s,s_k}$ and a tournament prestige parameter $\beta_{k}$. The surface similarity factor $\alpha_{s,s_k}$ down-weights matches played on surface $s_k$ when constructing the graph for target surface $s$, reflecting the reduced transferability of performance across surfaces, with $\alpha_{s,s} = 1$ for same-surface matches. The tournament prestige factor $\beta_{k}$ assigns weights based on match $k$'s tournament tier relative to Grand Slams, acknowledging that results from higher-prestige events may be more indicative of true player dominance.

These parameter values are optimised using the Non-dominated Sorting Genetic Algorithm II (NSGA-II), described in Section~\ref{subsec:paramopt}. The surface-specific dominance score is updated for each new tournament snapshot $n$ between players $u$ and $v$ on surface $s$ as follows:
\begin{equation}
    D_n^s(u,v) = \frac{\sum_{k} \alpha_{s,s_k} \beta_k \phi_k g_k(u,v)}{\sum_{k} \alpha_{s,s_k} \beta_k \phi_k},\label{eq:dominance_decay}
\end{equation}
where the sum ranges over all matches $k$ between players $u$ and $v$ that occurred before snapshot $n$, and $g_k(u,v)\in[0,1]$ is the proportion of games won by player~$u$ against $v$ in historical match~$k$, and $s_k$ is the surface of historical match $k$. $\alpha_{s,s_k}$ captures surface similarity, $\beta_k$ captures tournament prestige, and $\phi_k$ captures match recency. This calculation is performed independently for each of the three surfaces, yielding three separate graph representations. Since $g_k(u,v)$ records the proportion of games won rather than the match outcome alone, each match is worth much more than one binary observation (measured in Section~\ref{subsec:robustness}).

The dominance score calculation produces a value $D_n^s(u,v)$, from which we create a single directed edge pointing from the player with the higher dominance score to the weaker player. Specifically, if $D_n^s(u,v) > 0.5$, we assign a directed edge from player $u$ to player $v$ with weight $D_n^s(u,v)$. Conversely, if $D_n^s(u,v) < 0.5$, we assign a directed edge from player $v$ to player $u$ with weight $1 - D_n^s(u,v)$. If $D_n^s(u,v) = 0.5$ exactly, no edge is created. This ensures that each player pair contributes at most one directed edge to a graph, with the edge always pointing from the dominant player to the dominated player. Finally, we remove an edge when its corresponding players have not met for longer than a maximum edge age, set by the hyperparameter optimisation to 1.55 years (Table~\ref{tab:paramtuning}).

\begin{figure}[tb]
{
\centering
\includegraphics[width=1\textwidth]{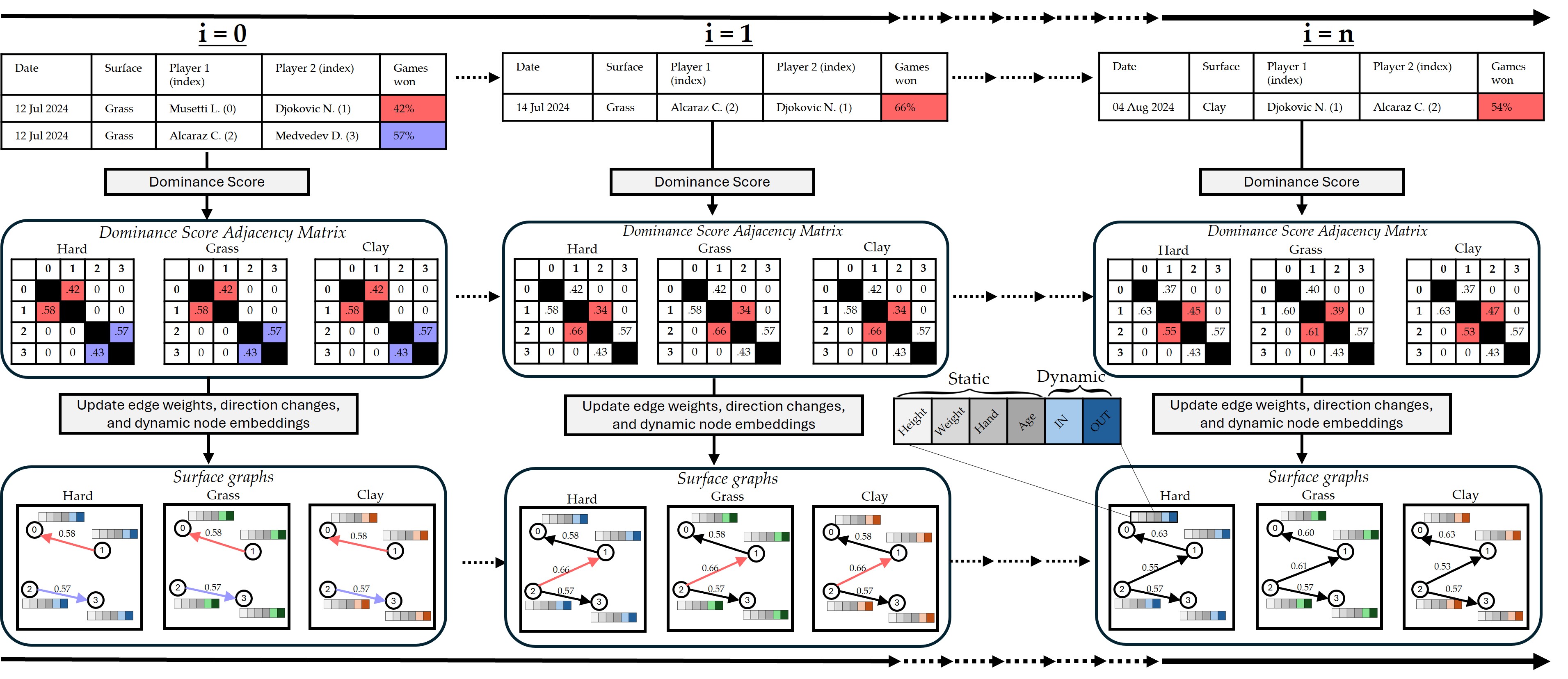}
\caption{Overview of the graph construction process. }
\label{fig:graph_construction}
}
{
\small Following each snapshot, the Dominance Score (Equation~\ref{eq:dominance_decay}) of the observed matchups is updated, determining the direction and weight of the edge between the players in the surface graphs $G^s_{i}$. Considering new graphs that are initialised at the 2024 Wimbledon semi-finals, $i=0$, two new edges are created. They point from the player who won $>50\%$ of games, towards the player with the lower percentage of games won. At $i=n$, Djokovic wins 54\% of the games in the Paris Olympics gold medal match, which weakens Alcaraz's dominance in all three surface graphs, most in the clay graph. Because of Alcaraz's stronger performance at Wimbledon, every Djokovic-Alcaraz edge still points towards Djokovic. Static node features are evaluated during parameter optimisation but excluded from the final model.
}
\end{figure}

This directed edge construction with asymmetric weights naturally preserves the intransitive relationships that exist between players. Figure~\ref{fig:graph_construction} provides an overview of our temporal graph construction process.

\subsection{MagNet}\label{subsec:magnet}
\noindent
To generate match outcome probability estimates, we use the spectral graph convolutional network (GCN) MagNet by \cite{zhang2021magnet}.\footnote{MagNet is accessible via the PyTorch Geometric Signed Directed library~\citep{he2024pytorch}.} We frame the problem as directional edge prediction, where for a scheduled match between players $u$ and $v$, MagNet estimates the edge direction (translating to set-win probability). We select MagNet specifically for its use of the magnetic Laplacian, a complex Hermitian matrix that enables learning from intransitive relationships rather than imposing transitive hierarchies. The magnetic Laplacian encodes undirected geometric structure (existence of head-to-head matches) in the magnitude of its entries and directional information (who dominated) in their phase.

MagNet uses a parameter $q \in [0, 0.25]$ that controls the extent to which directional information is incorporated. When $q=0$, the model disregards which player has historically dominated the head-to-head matchups, whereas at $q=0.25$, this dominance information is maximally integrated. We optimise $q$; the search selects $q=0.238$, near its maximum, where directional information is most strongly incorporated.

The model is trained to predict set outcomes by minimising cross-entropy loss with the Adam optimiser. The learning rate ($1.8\times10^{-3}$) is tuned jointly with the graph and architecture hyperparameters (Section~\ref{subsec:paramopt}), while weight decay ($10^{-5}$) and dropout ($0.3$) are fixed at standard defaults. Complete mathematical details are provided in \ref{app:magnet}.

In our dataset, there are both best-of-5 matches (played at men's Grand Slams) and best-of-3 matches (all women's matches and non-Grand Slam men's matches). In a best-of-3 match, there is a greater chance for the lower-ranked player to win as there are fewer opportunities for the stronger player to assert their dominance. Therefore, when producing probability estimates for the out-of-sample test set, we calculate match outcome probabilities for a best-of-3 sets match $\hat{P}_3$ and a best-of-5 sets match $\hat{P}_5$ under the assumption of independent and identically distributed sets:

\begin{equation}
    \hat{P}_{3}(u,v) = \hat{p}_{uv}^2 + 2\hat{p}_{uv}^2(1-\hat{p}_{uv}), \quad \hat{P}_{5}(u,v) = \hat{p}_{uv}^3 + 3\hat{p}_{uv}^3(1-\hat{p}_{uv}) + 6\hat{p}_{uv}^3(1-\hat{p}_{uv})^2
\end{equation}

\noindent where $\hat{p}_{uv}$ is the set-win probability for player $u$ against player $v$ output by MagNet. 

\subsection{Walk-forward validation}\label{subsec:walkfwdval}
\begin{figure}[tb]
\centering
\includegraphics[width=0.75\textwidth]{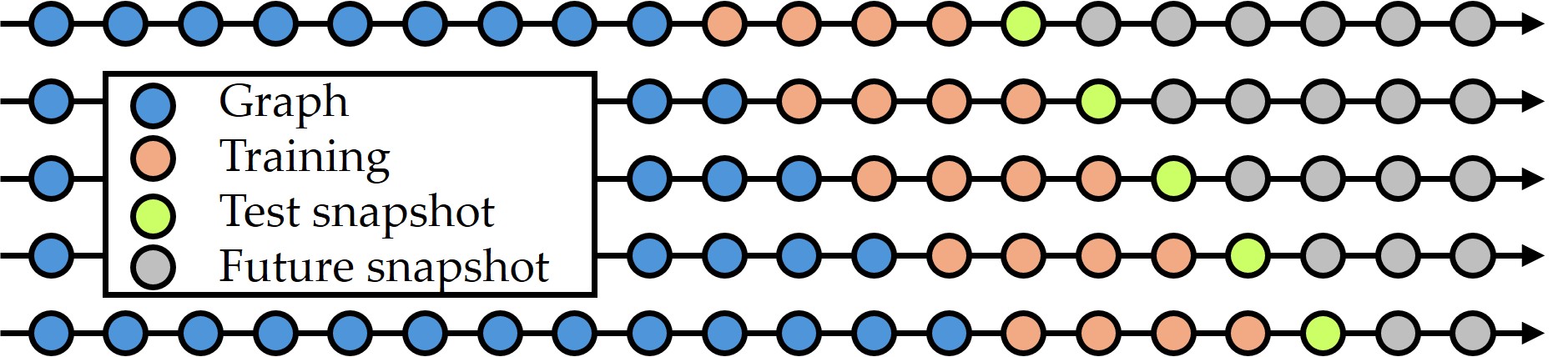}
\caption{The walk-forward validation process. After each pass, predictions are generated and the model is retrained; the historical graph grows as matches are added. For illustration, the schematic is not to scale: in practice, the historical graph and sliding training window span several hundred snapshots.}
\label{fig:walk_forward_validation}
\end{figure}
\noindent
With the temporal directed graph representation of tennis matches described in Section~\ref{subsec:graphrepresentation}, we adopt a walk-forward validation approach, shown in Figure~\ref{fig:walk_forward_validation}, to optimise parameters on a validation set and evaluate the model on an out-of-sample test set. 

At each snapshot, three surface graphs are constructed from the earliest 85\% of historical matches. The model is then trained to predict the most recent 15\% of matches, held out as labelled edges, from the graph built on the earlier matches.

The MagNet model parameters are optimised by training on this 15\% edge set. The model is initially trained for 100 epochs. To adapt to evolving player dynamics, it is then retrained for an additional 30 epochs every 38 snapshots, which we approximate to a quarter of a year. Each training run completes in under 10 seconds on consumer-grade hardware (RTX 2070 Super, Ryzen 3600). 

Following each training run, the model predicts outcomes for the next chronological snapshot. Once the true outcomes for a given snapshot are observed, its matches are integrated into the historical graph. The fixed-size training window then advances, redefining the 15\% training set to include the most recent matches for the subsequent cycle.

\subsection{Parameter optimisation}\label{subsec:paramopt}
\noindent
To identify the optimal parameters for graph construction, model architecture, and training, we conduct 250 trials using walk-forward validation. Each tour's match history is split chronologically: an initial window seeds the dominance graph, after which the model trains and hyperparameters are scored on a validation period ending in early 2022 ($8{,}435$ matches across both tours). All later matches, through October 2025, form the held-out test set of $12{,}052$ matches (Section~\ref{sec:results}) and are not used in tuning.

We select the final configuration in three stages: an initial optimisation over the full hyperparameter set, an ablation that discards the components which do not improve the validation Brier score (Section~\ref{subsec:ablation}), and a re-tuning of the reduced set. The configuration reported in this section is that of the final re-tuning, which optimises the 12 hyperparameters retained after ablation.

\begin{table}[tb]
    \centering
    \caption{Search spaces and selected values for the graph-construction and architecture hyperparameters.}
    \label{tab:paramtuning}
    \small
    \begin{tabular}{@{}lcc@{}}
    \toprule
    \textbf{Parameter} & \textbf{Search Space} & \textbf{Optimal Value} \\
    \midrule
    \multicolumn{3}{@{}l}{\textit{Model Architecture}} \\
    $K$, Chebyshev filter order & 1, 2, 3 & 1 \\
    Hidden units & 16, 32, 64, 128 & 16 \\
    Number of layers & 1, 2 & 2 \\
    Use activation function & True, False & False \\
    Directional weighting $q$ & $0.10$--$0.25$ & 0.24 \\
    \midrule
    \multicolumn{3}{@{}l}{\textit{Dominance signal}} \\
    Granularity & games, sets, points & games \\
    Min.\ dominance percentile & $0.0$--$0.5$ & 0.10 \\
    \addlinespace
    \multicolumn{3}{@{}l}{\textit{Time decay}} \\
    $\lambda$ & $0.0$--$2.5$ & 0.34 \\
    Max.\ edge age & $0.0$--$8.0$ & 1.55 \\
    \addlinespace
    \multicolumn{3}{@{}l}{\textit{Surface transferability (symmetric)}} \\
    $\alpha_{\text{h,c}}$ & $0.0$--$0.5$ & 0.41 \\
    $\alpha_{\text{h,g}}$ & $0.0$--$0.5$ & 0.22 \\
    $\alpha_{\text{c,g}}$ & $0.0$--$0.5$ & 0.40 \\
    \bottomrule
    \end{tabular}
    \begin{tablenotes}
    \item[*] \hspace{-2.5pt} Notes. Tuned parameters were optimised under uniform distributions over the stated ranges; comma-separated values denote discrete choices. Surface transfer is symmetric ($\alpha_{s,s_k}=\alpha_{s_k,s}$), with subscripts denoting surfaces: hard (h), clay (c), grass (g). In line with the ablation, tournament tiers are weighted uniformly ($\beta=1$).
    \end{tablenotes}
\end{table}

Table~\ref{tab:paramtuning} summarises the search space used for each parameter during optimisation, along with the optimal values found by applying the NSGA-II algorithm by \cite{deb2002fast}.\footnote{We use the Python library Optuna \citep{akiba2019optuna} to implement the NSGA-II algorithm.} We adopt a multi-objective framework, treating the Brier scores for men's and women's tours as separate objectives. We identify Pareto-optimal solutions that have a favourable trade-off between men's and women's performance, selecting a unified parameter set that balances predictive accuracy across both genders. The final model has 12 graph-construction and architecture hyperparameters, listed in Table~\ref{tab:paramtuning} and retained after the ablation in Section~\ref{subsec:ablation}, all tuned jointly with the Adam learning rate.

Given the model's complexity relative to traditional sports forecasting approaches, we select conservative search bounds for the graph construction parameters, described in Section~\ref{subsec:graphrepresentation}. Tournament tiers are weighted uniformly, as tier-specific weights did not improve predictive accuracy in our ablation.  

For the model architecture, we optimise several hyperparameters. The most important of these are the order of the Chebyshev polynomial approximation in the spectral graph convolution and the number of complex graph convolutional layers, collectively determining the size of the local neighbourhood from which player information is aggregated. We also optimise the dimensionality of the hidden channels, which controls the model's capacity to learn complex features. Furthermore, we evaluate the inclusion of a complex-valued ReLU activation function, which introduces non-linearity and allows the model to capture more intricate patterns.  

\begin{figure}[tb]
    \centering
    \includegraphics[width=0.75\textwidth]{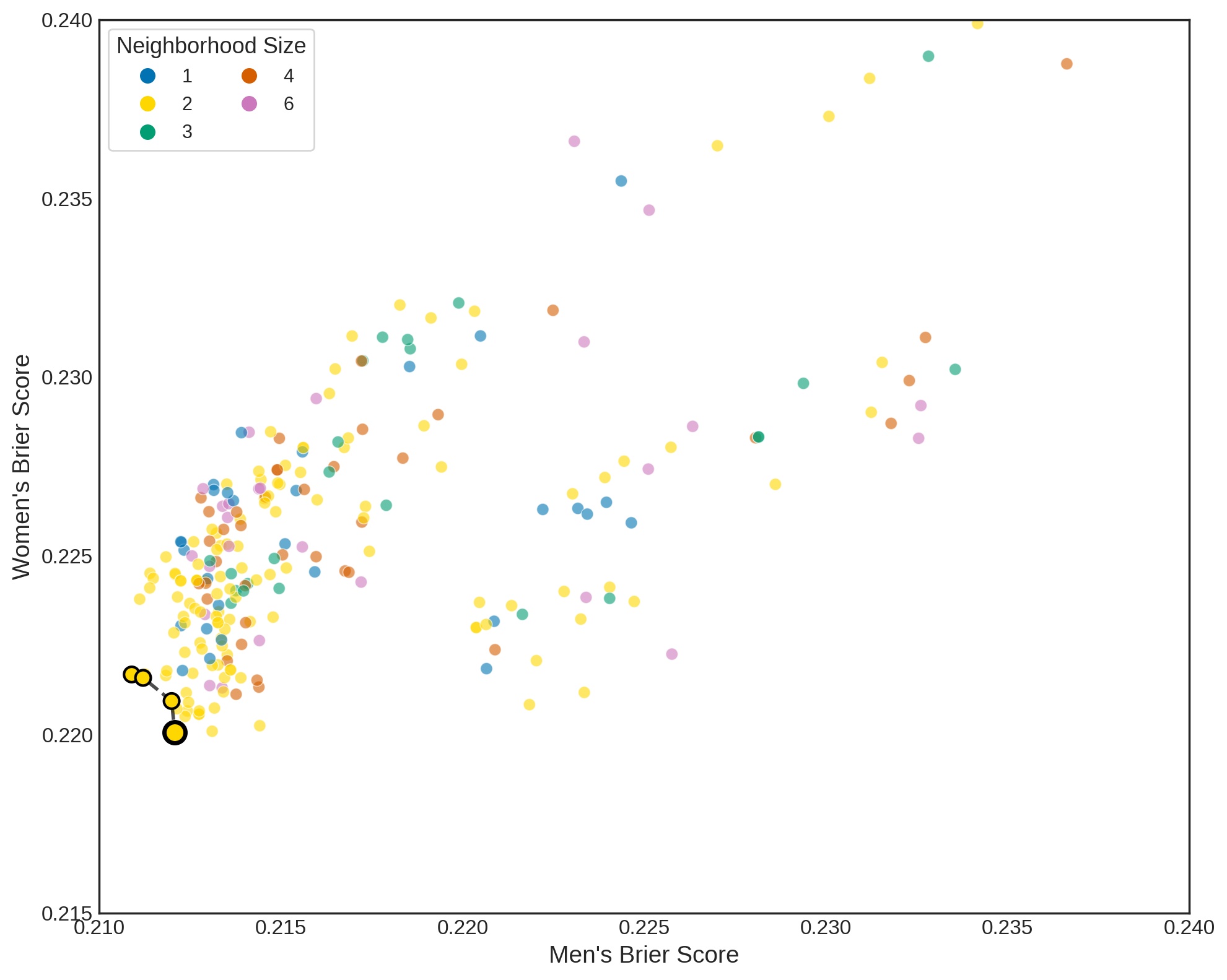}
    \caption{Brier scores from 250 parameter optimisation trials, with Pareto-optimal trials highlighted. Colour indicates receptive field size ($K \times L$ hops).}
    \label{fig:pareto_front}
\end{figure}

From the Pareto-optimal solutions shown in Figure~\ref{fig:pareto_front}, we select the trial with $K=1$ and $L=2$ layers and therefore an effective receptive field spanning $2$ hops. This neighbourhood size captures direct opponents and their immediate neighbours, providing sufficient context for detecting intransitive cycles while avoiding overfitting to global network properties that characterise existing graph-theoretic measures like eigenvector centrality.

The ablation that follows confirms the retained set is not redundant: removing any of its main components measurably degrades the validation Brier score.

\subsection{Ablation}\label{subsec:ablation}

\begin{table}[tb]
\centering
\small
\caption{Component ablation of the tuned configuration on the validation set: pooled Brier-score change for each variant.}
\label{tab:ablation_loo}
\begin{tabular}{@{}lr@{}}
\toprule
Variant & $\Delta$Brier \\
\midrule
\multicolumn{2}{@{}l}{\textit{Remove a component}} \\
Dominance graph (entire)                  & $+0.0350$ \\
Edge-age decay                            & $+0.0118$ \\
Multi-hop receptive field ($L{=}2$)       & $+0.0095$ \\
In/out-degree node features               & $+0.0058$ \\
Cross-surface edges                       & $+0.0043$ \\
Dominance edge weights                    & $+0.0013$ \\
Directional weighting ($q$)               & $+0.0007$ \\
\midrule
\multicolumn{2}{@{}l}{\textit{Simplify or add}} \\
Static player attributes (add)            & $+0.0005$ \\
Uniform tournament-tier weights           & $-0.0001$ \\
Symmetric surface transfer                & $0.0000$ \\
Games-level dominance signal              & $0.0000$ \\
\bottomrule
\end{tabular}

\vspace{1mm}
{\footnotesize Entries are each variant's Brier score (pooled across both tours by match count) minus the full-model reference of $0.2142$. Positive values indicate a degraded forecast.}
\end{table}

\noindent
We assess each component's contribution with a component ablation on the validation set, altering the full tuned configuration of the selection stage (Section~\ref{subsec:paramopt}) one component at a time: removing each retained component, and applying each simplification or addition considered for the final model (Table~\ref{tab:ablation_loo}).

Two findings stand out. First, the predictive signal is carried overwhelmingly by the graph structure. Removing the dominance graph entirely (leaving self-loops only, with fixed random node features) raises the pooled Brier score from $0.2142$ to $0.2492$, and the most consequential individual components are graph-structural: edge-age decay ($0.0118$), the multi-hop receptive field ($0.0095$), and the in/out-degree node features ($0.0058$). By contrast, the directional weighting $q$ ($0.0007$) and the dominance edge weights ($0.0013$) are the two smallest effects: the explicit direction and magnitude of each individual matchup add little to overall accuracy, consistent with the model's value being the broad-based complementary signal of Section~\ref{subsec:encompassing} rather than local intransitivity decoding.

Second, descriptive node features beyond degree do not help. Adding the static player attributes (height, weight, date of birth, handedness) raises the Brier score by $0.0005$. The full model therefore retains only the graph-derived in/out-degree as node features, and these are themselves valuable: removing them is among the largest single-component effects.

The remaining choices are simplifications the ablation shows cost nothing. Symmetric surface transfer ($\alpha_{s,s_k} = \alpha_{s_k,s}$) and uniform tournament tiers each move the pooled Brier score by at most $0.0001$, so we adopt both, shrinking the hyperparameter set before the final re-tuning. The games-level dominance signal performs no worse than the set- or point-level alternatives, so it is the level we retain.

\section{General Predictive Performance}\label{sec:results}
\noindent
We evaluate the proposed graph-based model against the Weighted Elo approach of \cite{angelini2022weighted}, the Elo approach of \cite{FiveThirtyEightSerenaGOAT}, a Bradley-Terry model \citep{BradleyTerry}, and bookmaker odds from Pinnacle Sports. We initialise the Elo rating systems from 2012, such that ratings are sufficiently stabilised for the out-of-sample test period. 

For the Bradley-Terry baseline, player strengths are estimated over a rolling two-year window by regularised maximum likelihood, computed via iterative Luce spectral ranking \citep{maystre2015fast}. For the bookmaker benchmark, we derive margin-adjusted implied probabilities from Pinnacle Sports odds using the model of \citet{shin1993}, which \citet{strumbelj2014} finds more accurate than basic margin normalisation.

The out-of-sample test period spans early 2022 to October 2025. We report two standard metrics. Classification accuracy is the proportion of matches whose winner is correctly identified (the predicted winner being the player assigned $\hat{p}_m > 0.5$). The Brier score \citep{brier1950verification} is the mean squared forecast error, $\frac{1}{N}\sum_{m=1}^{N}(\hat{p}_m - o_m)^2$, with $o_m \in \{0,1\}$ the outcome of match $m$; it rewards accuracy and calibration jointly. Lower Brier score and higher accuracy indicate better forecasts.

\begin{table}[tb]
    \centering
    \caption{Model Performance vs Baselines: Men's and Women's Results by Surface.}
    \label{tab:performance_summary}
    \resizebox{1.0\linewidth}{!}{%
    \begin{threeparttable}
    \begin{tabular}{l l r *{5}{r r}}
    \toprule
            &        &       & \multicolumn{2}{c}{\bf{Model}} & \multicolumn{2}{c}{Elo} & \multicolumn{2}{c}{WElo} & \multicolumn{2}{c}{BT} & \multicolumn{2}{c}{PS} \\
    \cmidrule(lr){4-5} \cmidrule(lr){6-7} \cmidrule(lr){8-9} \cmidrule(lr){10-11} \cmidrule(lr){12-13}
    Gender & Surface & {Count} & {Acc} & {Brier} & {Acc} & {Brier} & {Acc} & {Brier} & {Acc} & {Brier} & {Acc} & {Brier} \\
    \midrule
    Men    & Clay   & 1710  & \textbf{0.667} & \textbf{0.209}   & 0.643 & 0.219   & \underline{0.653} & \underline{0.215}   & 0.637 & 0.217   & 0.708 & 0.190 \\
    Men    & Grass  & 741   & 0.676 & \underline{0.205}   & \underline{0.680} & \underline{0.205}   & \textbf{0.687} & \textbf{0.202}   & 0.649 & 0.212   & 0.714 & 0.185 \\
    Men    & Hard   & 3604  & 0.656 & \textbf{0.214}   & \underline{0.659} & 0.215   & \textbf{0.661} & \textbf{0.214}   & \underline{0.659} & 0.217   & 0.683 & 0.199 \\
    \midrule
    Women  & Clay   & 1539  & 0.663 & \underline{0.210}   & \underline{0.668} & 0.211   & \textbf{0.669} & \textbf{0.209}   & 0.650 & 0.215   & 0.700 & 0.192 \\
    Women  & Grass  & 738   & 0.654 & \textbf{0.218}   & \underline{0.659} & 0.221   & \textbf{0.664} & \textbf{0.218}   & 0.649 & 0.223   & 0.663 & 0.207 \\
    Women  & Hard   & 3720  & \underline{0.646} & \underline{0.219}   & 0.641 & 0.221   & \textbf{0.647} & \textbf{0.218}   & 0.634 & 0.223   & 0.675 & 0.205 \\
    \midrule
    Both   & All    & 12052 & \underline{0.657} & \textbf{0.214}   & 0.653 & 0.217   & \textbf{0.658} & \textbf{0.214}   & 0.646 & 0.219   & 0.687 & 0.198 \\
    \bottomrule
    \end{tabular}
    \begin{tablenotes}
      \item[*] \hspace{-2.5pt} Notes. \textbf{Bold} indicates best performance and \underline{underlined} indicates second-best among the evaluated models (excluding PS benchmark) for each surface/gender combination, with values tied at three decimal places sharing a mark. Count refers to the number of matches in the out-of-sample test set. All metrics rounded to three decimal places. Elo represents the standard Elo rating system by \cite{FiveThirtyEightSerenaGOAT}. WElo represents the Elo extension by \cite{angelini2022weighted}. BT is the Bradley-Terry model of \cite{BradleyTerry}. PS represents Pinnacle Sports odds with \cite{shin1993} margin adjustment.
    \end{tablenotes}
    \end{threeparttable}
    }
\end{table}

Table~\ref{tab:performance_summary} reports performance overall and by surface. Our model is competitive with the rating-system baselines: it ties Weighted Elo \citep{angelini2022weighted} on Brier score (0.214), with Weighted Elo marginally ahead on accuracy (65.8\% vs 65.7\%); standard Elo (65.3\%, 0.217) and the Bradley-Terry model (64.6\%, 0.219) follow.

The bookmaker odds' implied probability, denoted as ``PS'', remains notably superior overall with 68.7\% accuracy and 0.198 Brier score, owing to its incorporation of expert knowledge, real-time market adjustments based on betting activity, and state-of-the-art models.

\subsection{Model calibration}\label{subsec:calibration}
\begin{figure}[tb]
    \centering
    \includegraphics[width=0.8\textwidth]{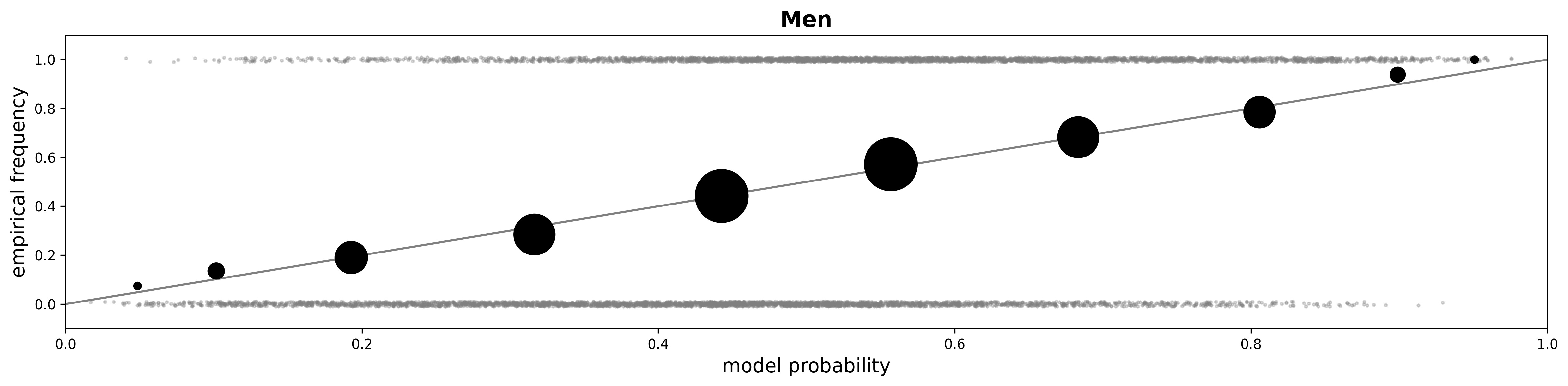}
    \includegraphics[width=0.8\textwidth]{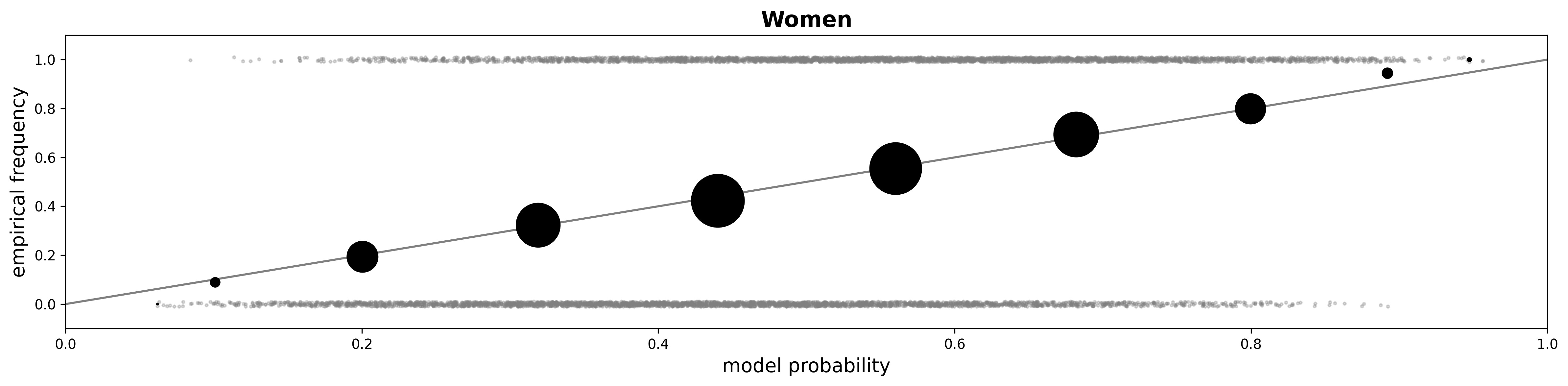}
    \caption{Calibration curves for the men's and women's predictions, showing predicted probabilities versus observed frequencies. Perfect calibration would align with the diagonal $y=x$ line.}
    \label{fig:calibration_curve}
\end{figure}
\noindent
To assess reliability, we follow \citet{BoshnakovKharratMcHale2017} and use recursive binning to plot calibration curves of the empirical outcome frequency $P(y = 1 \mid q)$ against predicted win probability $q$. 
Figure~\ref{fig:calibration_curve} shows predicted probabilities closely tracking observed frequencies for both tours, indicating strong calibration. The bias-corrected Murphy decomposition \citep{murphy1973new, ferro2012bias} of the full test-set Brier score quantifies this: the model's reliability term is below $0.0001$, against $0.0012$ for WElo, while WElo retains slightly higher resolution ($0.036$ vs $0.035$). These offsetting terms leave the two near-identical in Brier score, with the model better calibrated and WElo slightly more discriminating.

\section{Intransitivity and Forecast Complementarity}\label{sec:discussion}
\noindent
A prior application of MagNet~\ifnum\BLIND=0{\citep{clegg2025gnn} }\else{[Anonymous, 2025a] }\fi used an unoptimised configuration with substantial methodological limitations, including poorly calibrated time decay parameters, lack of cross-surface information flow, and reduced tournament coverage (detailed comparisons in \ref{app:mathsportext}). The present work addresses these limitations through systematic tuning and a complete cross-surface graph construction.

We investigate whether MagNet's advantages stem specifically from its ability to handle intransitive relationships.
Traditional rating systems cannot capture such structures because they rely on scalar player strength representations, assigning each athlete a single power rating (or perhaps separate serve and return point-win probabilities, as in the point-based model of \cite{o2008probability}).

\begin{figure}[tb]
    \centering
    \includegraphics[width=0.35\textwidth]{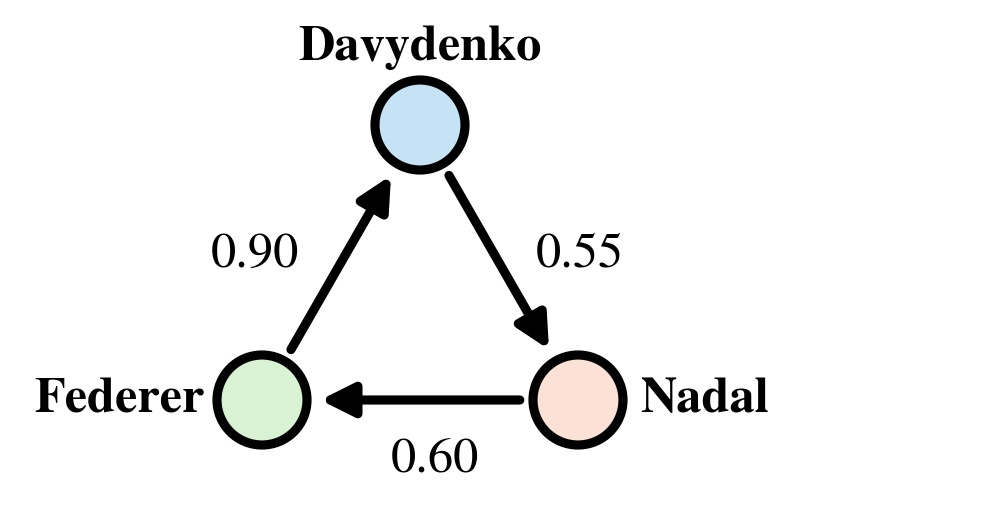}
    \caption{An intransitive triangle among Federer, Nadal, and Davydenko. Edge weights are career head-to-head match-win proportions. Each directed edge indicates a consistent head-to-head advantage.}
    \label{fig:non-transitive_triangle}
\end{figure}

This limitation is particularly relevant in tennis, where stylistic matchups can create intransitive relationships that our graph-based approach is designed to utilise. A well-known example of an intransitive relationship in tennis is shown in Figure~\ref{fig:non-transitive_triangle} and occurred during Roger Federer's peak in the mid-to-late 2000s, involving Federer, Rafael Nadal, and Nikolay Davydenko. Although Federer was the dominant world number one, Nadal consistently defeated him by exploiting his one-handed backhand with heavy, high-topspin forehands. In turn, Nadal was often troubled by Davydenko (when not playing on clay courts), whose flat, early baseline hitting neutralised Nadal's spin and pace, resulting in Davydenko's repeated success on hard courts. Completing the intransitive loop, Federer, despite his struggles against Nadal, dominated Davydenko, using variety and slice to disrupt Davydenko's rhythm.

\subsection{Quantifying intransitivity}\label{subsec:intransitivitymeasurement}
\noindent
We quantify intransitivity primarily with a model-free, ordinal count of cyclic head-to-head triads, in the tradition of circular-triad statistics \citep{kendall1940method, shizuka2012social}. The raw-triad count $T_h(u,v)$ counts common opponents $c$ for which the directed-majority wins among the three pairs $\{(u,c),(v,c),(u,v)\}$ form a 3-cycle, with each pair supported by at least $h$ prior head-to-head matches (default $h=1$). Pairs that have never met contribute no triad. Being an integer count of raw results, $T_h$ inherits neither the model's dominance scores nor the instability of a logit transform near certain outcomes, and it is the binning variable for every forecast-comparison test in this section (Sections~\ref{subsec:encompassing}--\ref{subsec:conflicts}).

\begin{figure}[tb]
\centering
\includegraphics[width=0.35\textwidth]{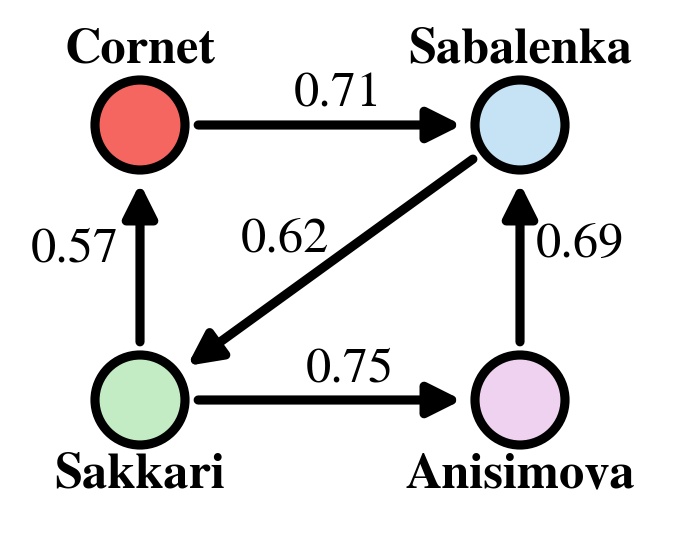}
\caption{Two intransitive cycles involving Maria Sakkari and Aryna Sabalenka, before their match in the 2021 Abu Dhabi WTA Women's Tennis Open. Each directed edge indicates a head-to-head dominance advantage.}
\label{fig:womens_nontransitive}
\end{figure}

By this count, cyclic dominance is common in professional tennis, and more so in the women's game: across the test set, women's matches contain 12.1\% more intransitive common-opponent triads than men's (13.87\% vs 12.37\% of triads are intransitive). Figure~\ref{fig:womens_nontransitive} illustrates two such cycles for one highly intransitive women's matchup.

We complement $T_h$ with a cardinal measure that reads intransitivity from the magnitudes of dominance rather than the direction of wins alone. Adapting the Hodge-decomposition measure of \citet{hamilton2024elo} to each match's common-opponent neighbourhood and scaling it by the matchup's accumulated evidence yields an evidence-weighted magnitude $I^*(A_{uv})$ (construction in \ref{app:cardinal}). Because $I^*$ derives from the model's dominance scores, it serves a descriptive role and enters none of the forecast-comparison tests. It is stable under shrinkage of those scores (\ref{app:shrinkage}) and agrees directionally with $T_h$ without being interchangeable.

\begin{figure}[tb]
    \centering
    \includegraphics[width=0.8\linewidth]{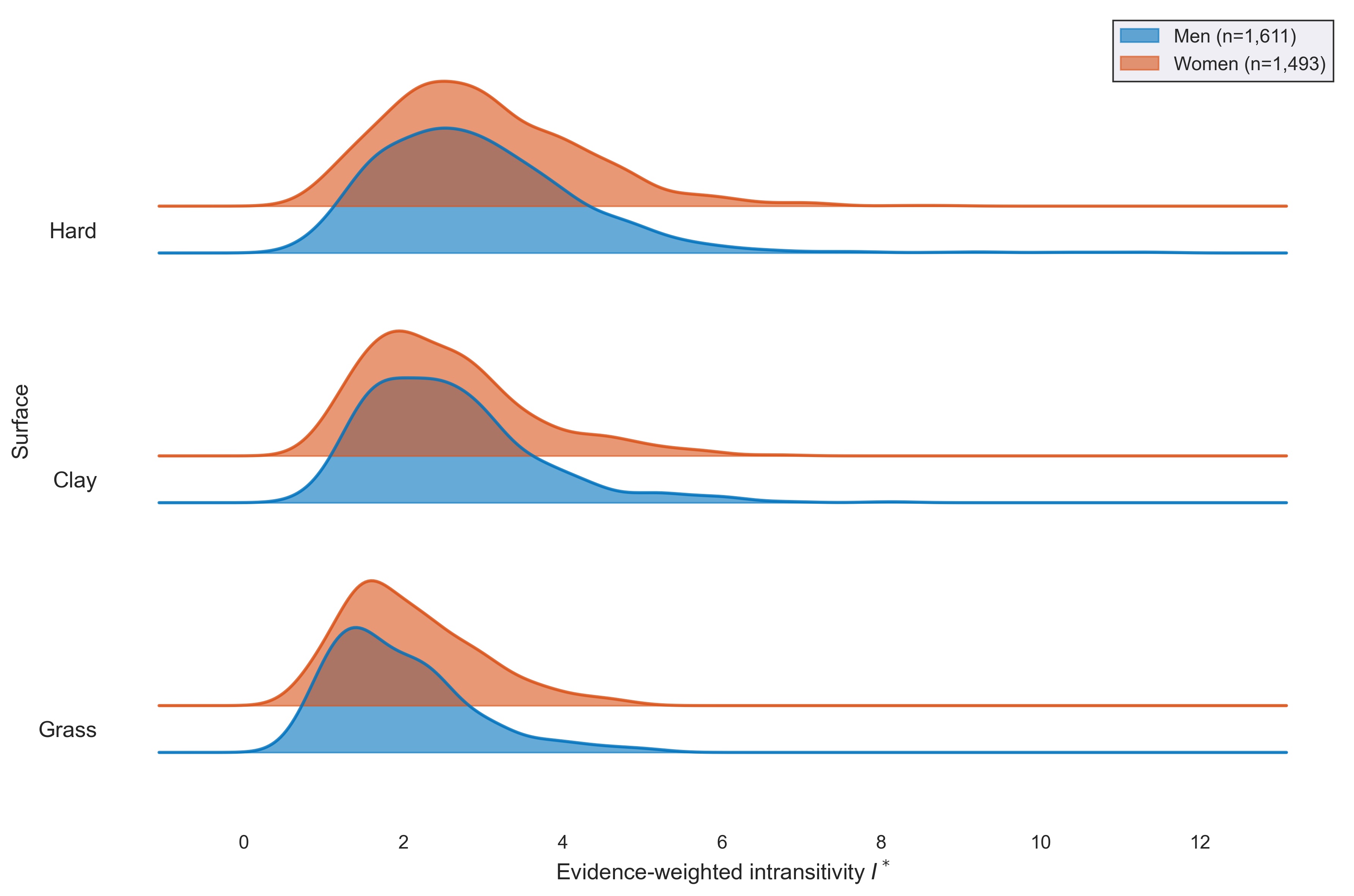}
    \caption{Distribution of evidence-weighted intransitivity $I^*(A_{uv})$ across the test set, by surface and gender, restricted to matches with $I^*>0$. Mean values: hard (men 2.94, women 2.99), clay (men 2.58, women 2.52), grass (men 1.95, women 2.09). Sample sizes: hard (men $n=1{,}007$, women $n=941$), clay (men $n=433$, women $n=387$), grass (men $n=171$, women $n=165$).}
    \label{fig:intransitivity_bias_by_surface}
\end{figure}

Figure~\ref{fig:intransitivity_bias_by_surface} shows distinct surface-specific patterns. Hard and clay courts exhibit right-skewed distributions, with most matchups at low-to-moderate $I^*$ and a long tail of upset-prone, highly intransitive matchups. The hard court's larger magnitudes reflect its greater match volume. Grass shows a concentration of moderate intransitivity despite low evidence, likely from its faster pace and lower friction. The gender gap in cyclic triads, clear under $T_h$, is muted under $I^*$ (2.77 vs 2.74 on the $I^*>0$ subset), because $I^*$ conditions on direct head-to-head evidence and 58.2\% of test matches fall between players who have not met.

\subsection{Model complementarity and encompassing}\label{subsec:encompassing}
\noindent
The measures of Section~\ref{subsec:intransitivitymeasurement} describe where intransitivity concentrates. Our model does not unconditionally outperform WElo (Section~\ref{sec:results}), so we test whether it nonetheless carries information orthogonal to WElo that improves a combined forecast. A forecast-encompassing test \citep{chong1986econometric} answers this directly. We refit the WElo logistic on the validation set:
\begin{equation}\label{eq:welo_refit}
\text{logit}\,\hat p_\text{W}(y) = a_0 + a_1\,\text{logit}\,\pi_\text{WElo}(y),
\end{equation}
where $\pi_\text{WElo}$ is the Weighted Elo win probability for match $y$. The test isolates our model's marginal contribution, and this Platt recalibration, with $\hat a_1 < 1$, shrinks an overconfident raw WElo and itself lowers the test-set Brier score from $0.2142$ to $0.2130$. The encompassing test incorporates our model's forecast $\pi_\text{G}$:
\begin{equation}\label{eq:combination}
\text{logit}\,\hat p_\text{W+G}(y) = b_0 + b_1\,\text{logit}\,\pi_\text{WElo}(y) + b_2\,\text{logit}\,\pi_\text{G}(y),
\end{equation}
following the combination form of \citet{granger1984improved}. Evaluating each logistic model on the test set, WElo encompasses our model if and only if $b_2 = 0$ in expectation. A one-sided Diebold-Mariano test \citep{diebold2002comparing} assesses whether the combination improves on recalibrated WElo's Brier score. Because the combination nests recalibrated WElo at $b_2 = 0$, with weights estimated once on validation and held fixed, the test is valid under the fixed-scheme asymptotics of \citet{giacomini2006tests}, which admit nested forecasts.

WElo is rejected as an encompassing forecaster on the test set: $B(\hat p_\text{W+G}) = 0.2113$ vs recalibrated WElo $B(\hat p_\text{W}) = 0.2130$ (DM $p<0.001$), with combination coefficient $\hat b_2 = +0.498$. Our model carries information not already present in WElo, and the signal is not confined to intransitive matches: evaluating the single validation-fitted combination on subsets of the test set, the gain is significant on both $T_1{=}0$ ($p < 0.001$) and $T_1{>}0$ ($p = 0.028$) matches.

Neither a surface-specific Elo (per-surface ratings) nor the Bradley-Terry model adds information to recalibrated WElo ($p = 0.28$ and $0.78$), so our model's signal is absent from every scalar rating we consider. In particular, it comes neither from surface specialisation nor from an alternative estimate of scalar strength. Its contribution also holds within each tour.

\begin{table}[tb]
\centering
\small
\caption{Correctness 2$\times$2 between our model and WElo on the test set.}
\label{tab:disagreement}
\begin{tabular}{@{}lrrr@{}}
\toprule
                & WElo right        & WElo wrong        & Total           \\
\midrule
Model right       & 7{,}036 & 877     & 7{,}913 \\
Model wrong       & 899     & 3{,}240 & 4{,}139 \\
\midrule
Total           & 7{,}935           & 4{,}117           & 12{,}052        \\
\bottomrule
\end{tabular}

\vspace{1mm}
{\footnotesize A forecast is correct when its favourite ($\hat p > 0.5$) wins. McNemar's test does not reject off-diagonal symmetry ($\chi^2 = 0.25$, $p = 0.62$).}
\end{table}

Table~\ref{tab:disagreement} quantifies how the two forecasts differ: exactly one is correct on 14.7\% of test matches, and these disagreements split almost evenly. McNemar's test does not reject symmetry, so neither systematically wins when the two disagree. They nonetheless err in complementary ways (Section~\ref{subsec:calibration}), so combining them extracts a signal absent from either alone.

\begin{table}[tb]
\centering
\small
\caption{Encompassing gain of the combination model across common-opponent subsets of the test set.}
\label{tab:gap_eff}
\begin{tabular}{@{}lrrrr@{}}
\toprule
Subset                          & $N$       & $\Delta B$ & $\eta$   & $p$-value \\
\midrule
No common opponents             & 745       & 0.0044     & 14.6\%   & 0.002 \\
Common opponents, no triad      & 6{,}275   & 0.0017     & 10.5\%   & $<$0.001 \\
Transitive triads               & 911       & 0.0004     & 8.9\%    & 0.371 \\
Intransitive triads             & 4{,}121   & 0.0012     & 10.7\%   & 0.028 \\
\bottomrule
\end{tabular}

\vspace{1mm}
{\footnotesize $N$ is the number of test matches in the subset; $\Delta B$ is the combination's Brier-score improvement over recalibrated WElo; $\eta$ (Equation~\ref{eq:gap_fraction}) is the share of the gap between recalibrated WElo and Pinnacle that it closes; $p$-values are from a one-sided Diebold-Mariano test of $B(\text{W+G}) < B(\text{W})$ on each subset.}
\end{table}

To gauge how much of the available improvement this represents, we take the Pinnacle closing odds as a sharp market forecast and measure complementarity efficiency as the share of the WElo-to-market headroom that our model closes:
\begin{equation}\label{eq:gap_fraction}
\eta = \frac{B(\text{W}) - B(\text{W+G})}{B(\text{W}) - B(\text{Pin})}.
\end{equation}
Table~\ref{tab:gap_eff} partitions the test set by local common-opponent structure. The gain is significant in three of the four cells, including the 745 matches whose players share no common opponents at all, where both the headroom and the share of it closed are largest ($\eta = 14.6\%$). On intransitive matches the model closes $\eta = 10.7\%$ of a smaller headroom, with the point estimate rising to $20.1\%$ ($h{=}2$) and $27.2\%$ ($h{=}3$) under the stricter triad-evidence thresholds of Section~\ref{subsec:robustness}. The market itself encompasses our model: adding $\pi_\text{G}$ to recalibrated Pinnacle gives no Brier-score improvement on either intransitivity subset (DM $p \geq 0.47$), so our model carries no information beyond the market even on intransitive matches. Its complementarity is relative to scalar rating systems rather than the market. The gain appears even where no local triad evidence exists, suggesting it is due to the global propagation of information through the dominance graph rather than the decoding of intransitivity.

\subsection{Robustness of the orthogonal-signal finding}\label{subsec:robustness}
\begin{table}[tb]
\centering
\small
\caption{Distribution of head-to-head meeting counts.}
\label{tab:h2h_dist}
\begin{tabular}{@{}lrr@{}}
\toprule
         & \multicolumn{2}{c}{Share (\%)} \\
\cmidrule(lr){2-3}
Prior meetings & Pairs & Test matches \\
\midrule
0    & --   & 58.2 \\
1    & 67.4 & 20.3 \\
2    & 17.8 & 9.7  \\
3    & 7.4  & 4.9  \\
4    & 3.4  & 2.7  \\
5$+$ & 3.9  & 4.1  \\
\bottomrule
\end{tabular}

\vspace{1mm}
{\footnotesize \emph{Pairs}: $24{,}374$ unique pairs by total meetings (median 1, max 28), with no zero row as a pair appears only once it has met. \emph{Test matches}: the $12{,}052$ test-set matches, grouped by the pair's number of prior meetings.}
\end{table}

\noindent
The dominance score (Equation~\ref{eq:dominance_decay}) is built from sparse head-to-head records, so we assess its robustness to verify that the encompassing finding of Section~\ref{subsec:encompassing} does not rely on sampling noise. Table~\ref{tab:h2h_dist} reports the distribution: the median pair has met once, and 58.2\% of test predictions involve players with no prior meeting. Each match nonetheless contains many games rather than one binary outcome. To account for dependence among those games, we use the one-way analysis-of-variance estimator of the intraclass correlation \citep{ridout1999estimating}, treating each match between a pair as a cluster of game outcomes. This gives intra-match correlations of $\rho = 0.0075$ for men and $0.0333$ for women, comparable with the near-iid behaviour of tennis points \citep{klaassen2001points}. The larger correlation for women is potentially due to the weaker serve dominance reported in the same study, which leaves game outcomes less tied to the service-alternation pattern.

By the design effect of \citet{kish1965survey}, $1+(\bar{m}-1)\rho$ for clusters of mean size $\bar{m}$, a match of mean length effectively contains $23.0$ design-effect-equivalent games of the $27.6$ played (men) and $12.8$ of the $21.7$ (women). Even the shortest possible best-of-three match, a 6--0 6--0 score known as a ``double bagel'', contains $11$ and $9$ effective games for men and women, respectively. The estimate may be conservative, as it attributes all variation between a pair's matches, including changes in form and surface, to within-match correlation. The resulting effective-game counts should also be interpreted as approximations because service alternation means that games do not have identical win probabilities. Importantly, these counts apply only to dominance within one match, as measured by the proportion of games won. They do not show that this dominance persists across meetings, which a single meeting cannot establish.

\begin{table}[tb]
\centering
\small
\caption{Encompassing gain on intransitive and transitive matches as $h$ (minimum prior meetings per triad edge) increases.}
\label{tab:threshold_sens}
\begin{tabular}{@{}llrrrr@{}}
\toprule
$h$ & Triads & $N$ & $\Delta B$ & $\eta$ & $p$-value \\
\midrule
1 & Intransitive & 4{,}121 & 0.0012 & 10.7\% & 0.028 \\
  & Transitive   &   911   & 0.0004 &  8.9\% & 0.371 \\
\addlinespace
2 & Intransitive & 1{,}362 & 0.0027 & 20.1\% & 0.012 \\
  & Transitive   & 1{,}174 & 0.0009 & 11.6\% & 0.233 \\
\addlinespace
3 & Intransitive &   736   & 0.0031 & 27.2\% & 0.034 \\
  & Transitive   &   630   & 0.0026 & 19.2\% & 0.055 \\
\addlinespace
4 & Intransitive &   243   & 0.0055 & 42.7\% & 0.037 \\
  & Transitive   &   511   & 0.0022 & 19.8\% & 0.120 \\
\bottomrule
\end{tabular}

\vspace{1mm}
{\footnotesize $N$: matches per cell (transitive: at least one consistently ordered triad and no cycle at threshold $h$). $\Delta B$ and $\eta$ (Equation~\ref{eq:gap_fraction}): the combined forecast's Brier reduction over recalibrated WElo, and the share of the WElo-to-Pinnacle headroom it closes. $p$-values: one-sided Diebold--Mariano test of $B(\text{W+G})<B(\text{W})$.}
\end{table}

Since a single meeting is weak evidence of a directed dominance relation, we re-evaluate the encompassing test under stricter triad-evidence thresholds. Table~\ref{tab:threshold_sens} increases the triad-evidence threshold $h$: the number of intransitive matches shrinks, yet the encompassing gain stays significant at every $h$. The gain also rises in both $\Delta B$ and $\eta$, but the same rise appears on transitive triads, so it tracks head-to-head evidence density rather than cyclicity. The intransitive gain is consistently larger, though the intransitive-transitive difference is not significant at any $h$. We do not consider $h$ beyond four, where the intransitive set has fallen to 243 matches, as a stricter threshold would leave too few for a reliable test. The finding is likewise insensitive to whether triad edges are oriented by match-, set-, or game-win majority, which agree on the $T_h{>}0$ label for 93--96\% of matches.

Since edges depend on which matches were played, we test their stability by leaving each contributing match out in turn. The graphs entering the test period contain $10{,}173$ active surface-specific edges, giving $17{,}565$ leave-one-match-out deletions. Of these, $8{,}826$ (50.2\%) leave no directed edge, $7{,}903$ (45.0\%) retain the original direction, and $836$ (4.8\%) reverse it. Conditional on an edge surviving, the direction is retained in 90.4\% of cases. The high removal rate reflects the sparse recent record, as $8{,}793$ edges have only one meeting within the 1.55-year activity window. Edges that reverse also have smaller weights, averaging $|D_n^s(u,v)-0.5|=0.036$, compared with $0.113$ for other multi-meeting edges. A single deletion therefore rarely reverses an edge but removes it about half the time, so the graph's topology is sensitive to individual matches. Reversals are concentrated among near-neutral relationships.

Additionally, we test the stability of the intransitivity classifications by perturbing the matches played and their game scores. We bootstrap each pair's match history, resampling matches with replacement 1,000 times, then redraw the game share of each sampled match from a Beta--Binomial distribution centred on the observed share and using the estimated intra-match correlation $\rho$. No set structure is imposed, so the replay's winner is the player with the redrawn game majority. A 13--11 match has its winner reversed in 35\% of redraws for men and 38\% for women, while a match with the most common best-of-three scoreline (6--3 6--4 for men and 6--3 6--2 for women) is reversed in 13\% and 8\%, respectively. An intransitive match at $h=1$ retains its label in 91\% of bootstrap replicates, falling to 66\% at $h=4$ as fewer cyclic triads support each classification.

For the encompassing gain itself, each replicate instead redraws every historical match's winner once and rebuilds every classification, with the model's forecasts held fixed so that variation reflects reclassification alone. The intransitive--transitive gap is positive in 42\% of replicates at $h=1$ and 85\% at $h=4$, matching the unconfirmed difference above. The Brier reduction on intransitive matches, however, stays positive in every one of 1,000 replicates at $h=1$ (90\% perturbation interval $[0.0007, 0.0014]$, against $0.0012$ observed) and in at least 99.7\% of replicates at every threshold. Since the forecasts are held fixed, this tests the stability of the classifications and subgroup results, not the sensitivity of the model's forecasts to the graph itself.

\subsection{Locating the complementary signal}\label{subsec:conflicts}
\noindent
Section~\ref{subsec:encompassing} attributed the combination's gain to global propagation through the dominance graph rather than to the decoding of individual cycles, but did not test the latter directly. We test it here. If the model is successfully decoding local cycles, it should side with the head-to-head record against the scalar favourite when the two disagree. We define a conflict as a match in which the WElo favourite differs from the leader of the prior head-to-head record, excluding matches with no prior meeting or a tied record.

Because a scalar rating imposes a single order, it must contradict the head-to-head record on some intransitive matchups, and conflicts concentrate there: 98\% occur on intransitive matchups, and conflicts are some four times more common on intransitive than transitive matchups (30.6\% vs 7.5\%). This association survives controls for neighbourhood size, head-to-head depth, and tour ($p < 0.001$).

On conflicts, a head-to-head direction indicator added to the WElo recalibration of Equation~\ref{eq:welo_refit} is insignificant on both the validation and test sets: the record predicts the winner no better than WElo alone, and our model, like the market, discounts it. The encompassing gain also disappears on the conflict subset itself (DM $p = 0.40$). The complementary signal is therefore not local cycle-decoding but global propagation of dominance structure across the graph: it is present even between players with no common opponents and grows on the more intransitive matchups our construction targets (Table~\ref{tab:threshold_sens}).

\section{Conclusion}\label{sec:conclusion}
\noindent
In this paper, we addressed the phenomenon of intransitive dominance between professional tennis players by constructing temporal directed graphs that preserve their relationships and applying a spectral graph convolutional network that learns directly from cyclic structure.

Our optimised model predicts tennis match outcomes with strong out-of-sample performance (65.7\% accuracy, 0.214 Brier score) and is competitive with existing prediction methods such as Weighted Elo (65.8\%, 0.214). Although the model does not improve on this baseline in accuracy, a forecast-encompassing test shows that a combined forecast significantly improves on Weighted Elo alone, with a Brier score of 0.211, demonstrating that our model provides complementary information.
Furthermore, the complementary signal of our model remains significant on intransitive matchups, which embody the cyclic structure that transitive rating systems cannot represent, though its absolute magnitude is roughly uniform across match types. This complementary value survives stricter head-to-head evidence thresholds and controlled reclassification of match histories.

However, the observed complementarity has limits. The model's overall predictive performance still trails bookmaker odds-implied probabilities significantly (0.214 vs 0.198 Brier score), and the approach is less interpretable than simpler ratings such as WElo. Moreover, even on the matchups where it complements rating systems, the model adds nothing to the market: combining it with Pinnacle's odds yields no measurable Brier-score improvement, so its complementary value is realised against transitive rating systems rather than the market itself. Two further limitations concern the graph itself. First, most edges rest on a single meeting. The graph is therefore sensitive to individual matches, and one meeting cannot establish that a head-to-head relationship persists. Second, the resampling analysis holds the model's forecasts fixed, so their sensitivity to the graph is not assessed. 

Future research could explore several promising directions. Constructing a more accurate measure of player-matchup intransitivity could enable deeper investigations into how rating systems handle cyclic dominance. Since player interactions and match histories can be encoded as graph structures, a natural next step is to test whether graph-based methods carry similar complementary value in other individual sports with strong stylistic components, such as mixed martial arts. There is also scope for a more detailed graph representation, given that our approach loses information when summarising rich matchup histories into simple edge weights. A method that learns from multiple edge attributes would be useful, and richer node features may also help. Overall, this study has shown that graph neural networks can capture forecasting information that transitive rating systems discard, complementing established ratings rather than replacing them.

\section*{Acknowledgements}
\ifnum\BLIND=1
    \noindent
    The authors have no conflict of interest to declare.
\else{
    \noindent
    LC's PhD is supported by a studentship from the Engineering and Physical Sciences Research Council (EPSRC) Doctoral Training Partnership (DTP), grant number EP/W524414/1. JC is supported by the UK Research and Innovation (UKRI) Engineering and Physical Sciences Research Council (EPSRC), grant number EP/Y028392/1: AI for Collective Intelligence (AI4CI). The authors have no conflict of interest to declare.
    }
\fi    

\bibliographystyle{elsarticle-harv}
\ifnum\BBL=1
    
\else
    \bibliography{references}
\fi

\newpage
\appendix
\section{Mathematical Details of MagNet}\label{app:magnet}
\noindent
This appendix provides the complete mathematical formulation of the MagNet architecture used in Section~\ref{subsec:magnet}. For clarity, we omit the court surface and snapshot notation, but the following applies independently to each temporal surface-graph. We also follow standard notation where bold uppercase letters (e.g., $\boldsymbol{A}$) denote matrices and bold lowercase letters (e.g., $\boldsymbol{x}$) denote vectors.

Given a static directed graph $G = (V, E, \mathbf{X^V}, W)$, we denote its adjacency matrix as $\boldsymbol{A} \in \mathbb{R}^{|V| \times |V|}$, where $A(u,v) = w_{uv}$ if an edge $(u,v) \in E$ exists indicating player $u$ has historically dominated player $v$, and $0$ otherwise. First, we compute the symmetrised adjacency matrix $\boldsymbol{A}_{s}$ and the diagonal degree matrix $\boldsymbol{D}_{s}$, defined as 

\begin{equation}
A_{s}(u,v) = \frac{1}{2}(A(u,v) + A(v,u)), \quad D_{s}(u,u) = \sum_v A_{s}(u,v).
\end{equation}

To preserve the directional information representing head-to-head dominance between players, the phase encoding matrix $\boldsymbol{\Theta}^{(q)}$ is calculated as:

\begin{equation}
\Theta^{(q)}(u,v) = 2\pi q(A(u,v) - A(v,u))
\end{equation}
\noindent
where $q \in [0, 0.25]$ is a tunable hyperparameter that controls the extent to which directional information is incorporated. When $q=0$, the model disregards which player has historically dominated the head-to-head matchups, whereas at $q=0.25$, this dominance information is maximally integrated.

For improved numerical stability during training, we use the normalised formulation of the magnetic Laplacian, which is defined as:
\begin{equation}
\boldsymbol{L}_N^{(q)} = \boldsymbol{I} - \boldsymbol{D}_{s}^{-1/2}\boldsymbol{A}_{s}\boldsymbol{D}_{s}^{-1/2} \odot \exp(i\boldsymbol{\Theta}^{(q)})
\end{equation}
The normalised magnetic Laplacian is a complex Hermitian matrix, which guarantees real eigenvalues, thus resolving the challenges posed by asymmetric adjacency matrices. This complex-valued representation captures both the existence of head-to-head history between players in the magnitude of its entries and the directional dominance pattern in their phase. For player pairs with no prior matches, the corresponding entries remain zero. 

To address the computationally expensive eigendecomposition of the La\-pla\-cian, MagNet implements the spectral filtering approach based on Chebyshev polynomial approximation, as introduced by \cite{defferrard2016convolutional}. As the eigenvalues of the normalised magnetic Laplacian lie in the range \([0, 2]\), the Laplacian is first rescaled to have eigenvalues in the range \([-1, 1]\) for the Chebyshev approximation:

\begin{equation}
\tilde{\boldsymbol{\mathcal{L}}} = \frac{2}{\lambda_{\max}}\boldsymbol{L}_N^{(q)} - \boldsymbol{I}
\end{equation}
where $\lambda_{\max}$ is the largest eigenvalue of $\boldsymbol{L}_N^{(q)}$. The parameter $K$ in the Chebyshev approximation corresponds to the order of the polynomial, defining the filter's receptive field as the $K$-hop neighbourhood of each node. The resulting filters aggregate information from nodes within $K$ hops, considering both incoming and outgoing edges. The phase of the Laplacian's entries, which encodes edge direction, allows the filter to process information from these two neighbourhoods distinctly. The filters act on node features $\boldsymbol{x}$ as follows:

\begin{equation}\label{eq:filters}
\boldsymbol{Y}\boldsymbol{x} = \sum_{k=0}^{K} \theta_k T_k(\tilde{\boldsymbol{\mathcal{L}}})\boldsymbol{x}
\end{equation}
where $\theta_k$ are the learnable filter parameters and $T_k$ are the Chebyshev polynomials of the first kind.

These filters operate across multiple layers of the network. When $L$ layers are stacked, the effective receptive field expands to $L \times K$ hops, allowing the model to aggregate information from increasingly distant players in the network. The multi-layer operation follows:
\begin{equation}
\boldsymbol{x}_{j}^{(\ell)} = \sigma\left(\sum_{i=1}^{F_{\ell-1}} \boldsymbol{Y}_{ij}^{(\ell)}\boldsymbol{x}_{i}^{(\ell-1)}\right)
\end{equation}
\noindent
where $\boldsymbol{x}_{j}^{(\ell)}$ is the $j$-th feature channel at layer $\ell$, $\sigma$ is a complex ReLU activation function, $F_{\ell-1}$ is the number of input channels to layer $\ell$, and $\boldsymbol{Y}_{ij}^{(\ell)}$ represents the convolution matrices implementing the spectral filters between the $i$-th input channel and the $j$-th output channel. After the final convolutional layer, the complex-valued node embeddings for all players are collected in the matrix $\boldsymbol{X}^{(L)} \in \mathbb{C}^{|V| \times F_L}$. To obtain real-valued node representations \(\boldsymbol{h}_u, \boldsymbol{h}_v \in \mathbb{R}^{2F_L}\), the matrix is unwound into real-valued features by separating real and imaginary parts into a matrix $\mathbb{R}^{|V| \times 2F_L}$. Edge features are extracted by concatenating the unwound node embeddings for each player pair $(u,v)$, resulting in edge representation $\boldsymbol{e}_{uv} = [\boldsymbol{h}_u; \boldsymbol{h}_v] \in \mathbb{R}^{4F_L}$.

The final prediction layer applies a linear transformation followed by softmax activation to produce an order-dependent probability vector for a given player pair, representing the estimated probability associated with the directed edge from \(u\) to \(v\) (indicating player \(u\)'s dominance over \(v\)):

\begin{equation}
\hat{\mathbf{z}}_{uv} = \text{softmax}(\mathbf{W}[\mathbf{h}_u; \mathbf{h}_v] + \mathbf{b})
\end{equation}
\noindent
where $\mathbf{h}_u$ and $\mathbf{h}_v$ are the node embeddings from the final layer, $[\mathbf{h}_u; \mathbf{h}_v]$ represents their concatenation, and $\mathbf{W} \in \mathbb{R}^{2 \times 4F_L}$ and $\mathbf{b} \in \mathbb{R}^2$ are learnable weight and bias terms. 

To mitigate the impact of player ordering \((u, v)\) on probability estimates and to ensure a balanced prediction, we average the model outputs from both directed perspectives. The final set-win probability for player $u$, denoted $\hat{p}_{uv}$, is defined as:
\begin{equation}
\hat{p}_{uv} = \frac{1}{2} \left( (\hat{\mathbf{z}}_{uv})_1 + (\hat{\mathbf{z}}_{vu})_2 \right)
\label{eq:avg_prob}
\end{equation}
where $(\cdot)_1$ refers to the first component of the probability vector (i.e., the probability of the first player in the input pair winning) and $(\cdot)_2$ refers to the second. This averaging ensures that the model's predictions are consistent regardless of player order and that the resulting probabilities sum to one, i.e., $\hat{p}_{uv} + \hat{p}_{vu} = 1$.

\section{Extended Initial Model}\label{app:mathsportext}
\noindent
In earlier work \ifnum\BLIND=0
  \citep{clegg2025gnn}
\else
  [Anonymous, 2025b],
\fi we proposed an unoptimised temporal graph model with a less sophisticated graph structure for the same task of tennis forecasting. To evaluate the robustness of this unoptimised approach, we increased the out-of-sample dataset to cover the period 31 October 2023 to 8 June 2025. By incorporating 844 additional matches, the total out-of-sample test set increased in size to 1,918 professional matches across both the original period (31 October 2023 to 8 September 2024) and the extended period (9 September 2024 to 8 June 2025). The dataset differs from the one we use here in that it only considers men's matches and excludes any 500-point matches other than the Halle Open and the Queen's Club Championships.

The initial implementation of temporal graph neural networks for tennis forecasting also differs in several key aspects from the model presented in our current work. Specifically, the graph edge weights are constructed with a time decay parameter of $\lambda = 0.01$, and there is no cross-surface graph updating. This extension provides further justification for the proper hyperparameter tuning and improved graph structure in this work.

\begin{table}[tb]
    \centering
    \small
    \caption{Model Performance Comparison By Surface: Extended Dataset.}
    \label{tab:extended_performance}
    \begin{threeparttable}
    \begin{tabular}{l *{3}{r r}}
    \toprule
            & \multicolumn{2}{c}{Count} & \multicolumn{2}{c}{Accuracy} & \multicolumn{2}{c}{Brier} \\
    \cmidrule(lr){2-3} \cmidrule(lr){4-5} \cmidrule(lr){6-7}
    Dataset & {MS} & {E} & {MS} & {E} & {MS} & {E} \\
    \midrule
    Clay    & 314 & 681 & 0.675 & 0.649 & 0.216 & 0.226 \\
    Grass   & 169 & 169 & 0.710 & 0.710 & 0.209 & 0.209 \\
    Hard    & 591 & 1068 & 0.638 & 0.624 & 0.221 & 0.228 \\
    \midrule
    All     & 1074 & 1918 & 0.660 & 0.640 & 0.218 & 0.226 \\
    \bottomrule
    \end{tabular}
    \begin{tablenotes}
      \item[*] \hspace{-2.5pt} Notes. MS refers to results presented in \ifnum\BLIND=0
  \cite{clegg2025gnn}
\else
  [Anonymous, 2025b],
\fi E refers to results from using the same methodology but with an extended dataset. While the original results used a dataset spanning 31 October 2023 to 8 September 2024, the extended dataset covers 31 October 2023 to 8 June 2025. All metrics rounded to three decimal places.
    \end{tablenotes}
    \end{threeparttable}
\end{table}

In Table~\ref{tab:extended_performance}, the results from the original out-of-sample test set are shown along with results when including the 844 additional matches. The unoptimised model achieves an overall classification accuracy of 64.0\% and a Brier score of 0.226, representing a decline from the original model's performance of 66.0\% accuracy and 0.218 Brier score. This deterioration is consistent across most performance metrics and surfaces, with the exception of grass courts, since there were no additional matches on grass. The difference is most pronounced on clay courts, where accuracy drops from 67.5\% to 64.9\%, with hard court performance declining from 63.8\% to 62.4\%.

We attribute the performance deterioration to several methodological differences. First, the dramatically different time decay parameter ($\lambda = 0.01$ vs $\lambda = 0.34$) significantly alters how historical match information influences current predictions, causing the graph to retain excessive influence from outdated match results that may no longer reflect current player form.

Second, the exclusion of 500-point tournaments reduces the graph density and available training information, particularly affecting predictions of players who compete primarily at these lower-tier events. Finally, the restriction of cross-surface information flow prevents the model from consistently learning across the year. This is most pronounced with grass tournaments, which necessitated the selective inclusion in the original work of the Halle Open and the Queen's Club Championships as a partial remediation.

The performance of the initial model with the extended dataset highlights the importance of our improved approach, with systematic hyperparameter optimisation and complete graph construction.

\section{Cardinal Intransitivity Measure}\label{app:cardinal}
\noindent
Section~\ref{subsec:intransitivitymeasurement} pairs the model-free raw-triad count with a cardinal measure, the evidence-weighted intransitivity $I^*(A_{uv})$, whose construction we give here. It adapts the Hodge-decomposition measure of \citet{hamilton2024elo} to each match's common-opponent neighbourhood. For players $u$ and $v$, the common opponents are those who have faced both before snapshot $n$:
\begin{equation}
\mathcal{C}_{uv} = \{c \in V : (u,c) \in E_n \text{ and } (v,c) \in E_n\}.
\end{equation}
We extract the $(|\mathcal{C}_{uv}| + 2) \times (|\mathcal{C}_{uv}| + 2)$ subgraph $G_{uv}$ on $\{u, v\} \cup \mathcal{C}_{uv}$ and form the advantage matrix $A_{uv}$ by a logit transform,
\begin{equation}
A_{uv}[a,b] = \log\left(\frac{w_{ab}}{1 - w_{ab}}\right),
\end{equation}
where $w_{ab}$ is the dominance score of player $a$ over player $b$ within the neighbourhood, clipped $10^{-6}$ away from $0$ and $1$ so the transform cannot diverge. This maps bounded dominance scores $[0,1]$ to unbounded advantages $(-\infty, +\infty)$. Because $w_{ba} = 1 - w_{ab}$, the transform is antisymmetric ($A_{uv}[b,a] = -A_{uv}[a,b]$), as the Hodge decomposition requires. The decomposition separates $A_{uv}$ into a transitive component $\text{grad} \circ \text{div}(A_{uv})$ and a cyclic remainder. \citet{hamilton2024elo} measure intransitivity as the ratio
\begin{equation}
I(A_{uv}) = \frac{1 + \|A_{uv} - \text{grad} \circ \text{div}(A_{uv})\|_F}{1 + \|\text{grad} \circ \text{div}(A_{uv})\|_F},
\label{eq:hamilton_original}
\end{equation}
which equals one when the cyclic and transitive components have equal magnitude, with values above one marking predominantly cyclic neighbourhoods. Equation~\ref{eq:hamilton_original} scores a whole neighbourhood uniformly, whereas we need a matchup-specific quantity, so we scale it by the accumulated evidence weight for $(u,v)$:
\begin{equation}
I^*(A_{uv}) = I(A_{uv}) \cdot \sqrt{\sum_{k} \alpha_{s,s_k} \beta_k \phi_k},
\label{eq:weighted_intransitivity}
\end{equation}
where the sum ranges over all matches $k$ between $u$ and $v$ before snapshot $n$, matching the denominator of Equation~\ref{eq:dominance_decay}. The accumulated weight captures evidence from match recency, tournament prestige, and surface correlations, with square-root scaling moderating highly-weighted matchups. Because $I^*$ is built from the model's dominance scores, it is descriptive only and enters none of the forecast-comparison tests, which bin on the model-free $T_h$. Its stability under shrinkage of those scores is established in \ref{app:shrinkage}.

\section{Robustness of Dominance Scores}\label{app:shrinkage}
\noindent
A potential concern with the dominance score $D^s_n(u,v)$ (Equation~\ref{eq:dominance_decay}) is estimation error arising from sparse head-to-head records. We assess this with Bayesian shrinkage at two levels.

Firstly, we shrink each pair's pooled empirical game share towards $0.5$ using its Beta$(\alpha,\alpha)$ posterior mean, with prior strengths $\alpha \in \{1,3,5\}$. Across the $17{,}364$ pairwise head-to-head records observed before the test period, the Kendall $\tau$ between empirical and shrunken shares is $0.99$ at $\alpha=1$ and $0.96$ at $\alpha=5$. 

Secondly, we shrink $D^s_n(u,v)$ directly towards $0.5$ by adding $2\alpha$ pseudo-observations of an even game share. At $\alpha=1$, the original and shrunken rankings by dominance strength (distance from $0.5$) have Kendall $\tau=0.64$. Being symmetric, this shrinkage cannot reverse an edge, so sensitivity of edge direction to the underlying record is assessed by the leave-one-match-out analysis of Section~\ref{subsec:robustness}.

Because $I^*(A_{uv})$ is derived from the dominance scores, we also recompute it from the shrunken scores while retaining the observed evidence weights. Its ranking remains stable across all prior strengths (Spearman $\rho\geq0.94$, Kendall $\tau\geq0.79$), with no match changing its $I^*>0$ status, indicating that the descriptive results for $I^*(A_{uv})$ in Section~\ref{subsec:intransitivitymeasurement} are not artefacts of unstable estimates.

\end{document}